\documentclass{article}

\newcommand{\dsetName}{MedEvidence}
\newcommand{\dsetSize}{284}
\newcommand{\dsetReviews}{100}
\newcommand{\dsetIncludedRefs}{329  }
\newcommand{\dsetFulltextRefs}{114}
\newcommand{\modelCount}{24}
\newcommand{\nSpecialties}{10}

    \usepackage[preprint]{neurips_2025}

\usepackage{pifont}                  
\newcommand{\cmark}{\ding{51}}       
\newcommand{\xmark}{\ding{55}}       
\usepackage{float}
\usepackage[table]{xcolor}
\usepackage{amssymb} 
\definecolor{lightgreen}{HTML}{DFF0D8}
\definecolor{lightred}{HTML}{F2DEDE}
\usepackage{wrapfig} 
\usepackage[utf8]{inputenc} 
\usepackage[T1]{fontenc}    
\usepackage{hyperref}       
\usepackage{url}            
\usepackage{booktabs}       
\usepackage{amsfonts}       
\usepackage{nicefrac}       
\usepackage{microtype}      
\usepackage{xcolor}         
\usepackage{graphicx}
\usepackage{subcaption}
\usepackage{adjustbox}
\usepackage{listings}
\usepackage{enumitem}
\usepackage{breakurl}
\usepackage{boxedminipage}
\usepackage{tcolorbox}
\tcbuselibrary{listings,skins,breakable}
\usepackage{titletoc}
\usepackage{changepage}
\usepackage{placeins}
\newtcolorbox{QualExample}[2]{
  breakable,
  colback=white,
  colframe={#1},
  title={#2}
}

\title{Can Large Language Models Match the Conclusions of Systematic Reviews?} 

\author{%
Christopher Polzak*$^1$
\And
Alejandro Lozano*$^1$
\And
Min Woo Sun*$^1$
\AND
James Burgess$^1$
\And
Yuhui Zhang$^1$
\And
Kevin Wu$^1$
\And
Serena Yeung-Levy$^1$
\AND
$^1$Stanford University
}

\begin{document}

\maketitle

\begin{abstract}
  Systematic reviews (SR), in which experts summarize and analyze evidence across individual studies to provide insights on a specialized topic, are a cornerstone for evidence-based clinical decision-making, research, and policy. Given the exponential growth of scientific articles, there is growing interest in using large language models (LLMs) to automate SR generation. However, the ability of LLMs to critically assess evidence and reason across multiple documents to provide recommendations at the same proficiency  as domain experts  remains poorly characterized. We therefore ask:\textbf{ Can LLMs match the conclusions of systematic reviews written by clinical experts when given access to the same studies?} To explore this question, we present \dsetName{}, a benchmark pairing findings from 100 SRs with the studies they are based on. We  benchmark \modelCount{} LLMs on  MedEvidence, including reasoning, non-reasoning, medical specialist, and models across varying sizes (from 7B-700B). Through our systematic evaluation, 
    we find  that  reasoning does not necessarily improve  performance, larger models do not consistently yield greater gains, and knowledge-based fine-tuning degrades accuracy on MedEvidence. Instead, most models exhibit similar behavior: performance tends to degrade as token length increases, their responses show overconfidence, and, contrary to human experts, all models show a lack of scientific skepticism toward low-quality findings.
These results suggest that more work is still required before LLMs can reliably match the observations from expert-conducted SRs, even though these systems are already deployed and being used by clinicians.  We release our codebase\footnote{\scriptsize{\url{https://github.com/zy-f/med-evidence}}} and benchmark\footnote{\scriptsize{\url{https://huggingface.co/datasets/clcp/med-evidence}} \\ $^*$ Equal contributions: \{clcp, lozanoe, minwoos\}@stanford.edu} to the broader research community to further investigate  LLM-based SR systems.



\end{abstract}

\section{Introduction}
\label{sec: intro}

\begin{figure}
    \makebox[\textwidth][c]{\includegraphics[width=1.2\textwidth]{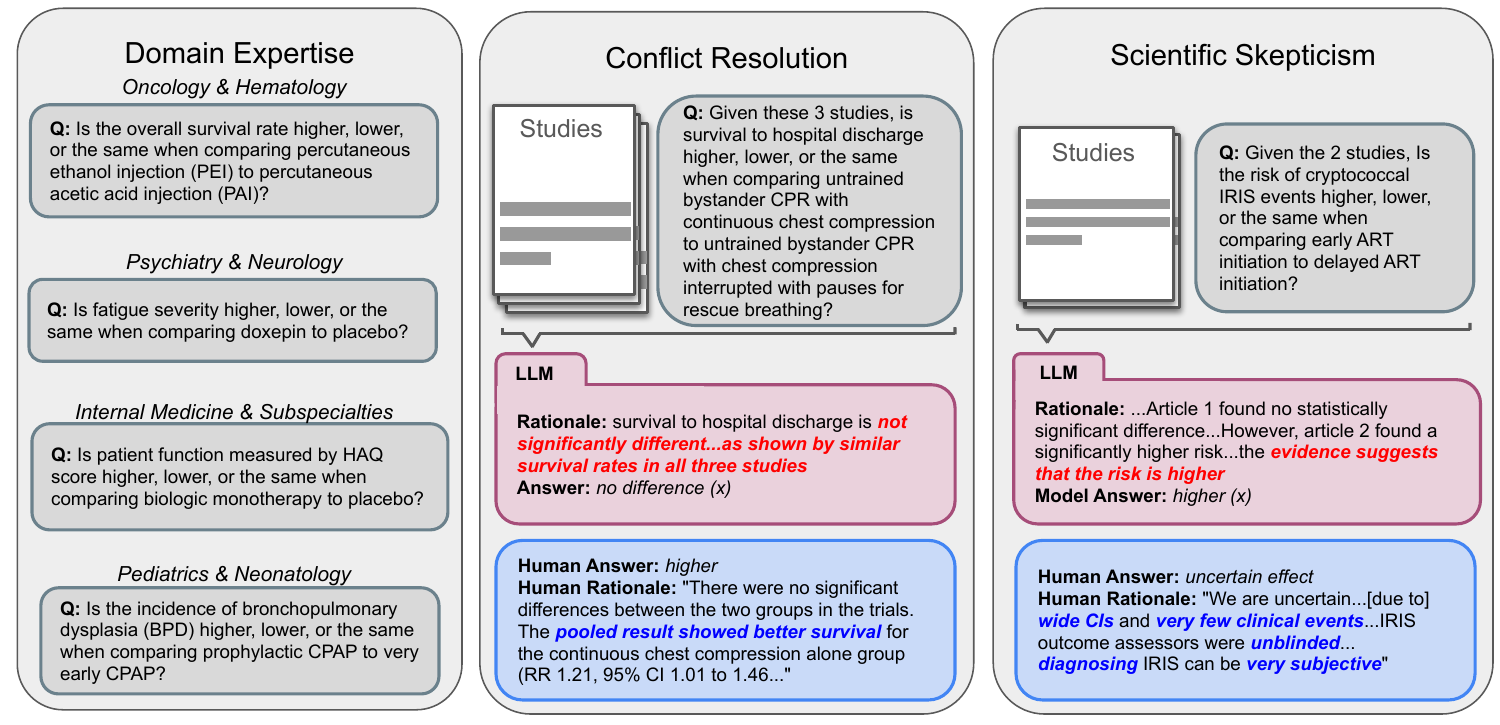}}
    \caption{Core skills evaluated by \dsetName{} including: medical domain expertise across 10 different specialties, synthesizing conflicting evidence, and applying scientific skepticism when studies exhibit a high risk of bias (e.g. due to small sample sizes or insufficient supporting evidence).} 
    \label{fig:pull_figure}
\end{figure}


As the number of published articles grows exponentially \cite{Bornmann2021}, manually synthesizing findings from multiple sources has become highly time-consuming. Thus, there is growing interest in developing automatic tools to process, synthesize, and extract insights from scientific literature \cite{lozano2023clinfo,scherbakov2024emergence}. In particular, large language model (LLM)-based systems could offer a promising solution for supporting and automating tasks such as conducting systematic reviews (SRs), which typically take an average of 67 weeks of intensive human effort \cite{fabiano2024optimize,riaz2024future}. For example, several LLM-assisted tools such as Deep Research \cite{openai2025deepresearch, google2025deepresearch}, Elicit \cite{elicit2025}, and Open Evidence \cite{openevidence2025}, have already been deployed and can be incorporated into the SR process to improve efficiency \cite{fabiano2024optimize}. The momentum behind these technologies is further exemplified by the U.S. Food and Drug Administration’s launch of an LLM-assisted scientific review pilot on May 2025 \cite{FDA2025}.

However, despite multiple deployments and efforts assessing scientific synthesis generation, the behavior of LLMs across key variables that influence generation remains poorly understood. In particular, their ability to synthesize findings from multiple studies—each varying in study type, population size, and risk of bias—and to navigate conflicting evidence (as medical findings may  contradict one another) is not well-characterized. Understanding these behaviors is essential, as medical knowledge is continually reshaped by new clinical trials, cohort studies, and expert opinions. Thus, like medical professionals do, LLMs must be capable of integrating the latest findings (e.g. via retrieval augmentation) \cite{ke2024development}, weighing the strength of varying evidence, and applying appropriate skepticism when needed to produce reliable, up-to-date recommendations (as shown in Figure \ref{fig:pull_figure}).

While prior work has successfully evaluated LLMs on their internal "static" medical knowledge \cite{lievin2024can,fleming2024medalign}, assessing LLMs' capability to reason across multiple sources and draw expert-level conclusions remains a significant challenge. Specifically, previous efforts have often evaluated LLMs' ability to generate summaries on a given topic. This approach requires a thorough review of every detail in the generated content and lacks easily verifiable ground truth; therefore, medical experts are typically needed to assess output accuracy \cite{NMA_caseStudy, SR_automation_endtoend_caseStudy, SR_automation_endtoend_caseStudy_webinar, riskOfBias_caseStudy,lozano2025biomedica}, making evaluation time-consuming and hard to scale. 
 To address this, we remove the complexity of evaluating long-format summaries and retrieving relevant papers to pose an even simpler, but fundamental question: \textbf{ Can LLMs replicate the individual conclusions of expert-written SRs when provided with the same source studies?} We explore this question in a controlled setting by collecting open-access SRs along with their associated reference articles.  We then extract individual findings and reformat them into a closed question-answering (QA) task to simplify evaluation. To this end, we introduce the following contributions:

\begin{itemize}[leftmargin=*]
    \item \textbf{\dsetName{} Benchmark} We introduce \dsetName{}, a human-curated benchmark of \dsetSize{} questions curated from the conclusions of \dsetReviews{} open-access SRs across \nSpecialties{} medical specialties. Each question evaluates comparative treatment effectiveness on clinical outcomes. All questions are manually transformed into closed-form question answering to enable large-scale evaluation. In addition, human annotators extract evidence quality (based on the SR’s analysis), determine whether full-text access is necessary, and collect the relevant sources needed to replicate the SR findings.
    

    \item \textbf{Large-scale evaluation on \dsetName{} } We leverage MedEvidence to perform an in-depth analysis of \modelCount{} LLMs spanning general-domain, medical-finetuned, and reasoning models. By utilizing MedEvidence's metadata, we dissect and examine success and failure modes, helping to identify targeted directions for future work.
\end{itemize}

\section{Related work}

\begin{table}[ht!]
\centering
\caption{Comparison of factuality and evidence reasoning benchmarks with medical focus. We compare MedEvidence to prior datasets across attributes relevant to systematic review-style reasoning. MedEvidence is the only dataset to satisfy all criteria.}
\label{tbl:related_work}
\resizebox{\textwidth}{!}{%
\begin{tabular}{lcccccccc}
\toprule
\shortstack{Dataset} & \shortstack{Size} & \shortstack{Topic} & Curation & \shortstack{Expert-Grounded \\ Answer} & \shortstack{Automated \\ Evaluation} & \shortstack{Multiple \\ Sources} & \shortstack{Evidence \\ Quality} & \shortstack{Source-Level\\ Concordance } \\
\midrule
Reason et al. & 4 & Medicine & Human & \cellcolor{lightgreen}\textcolor{black}{\cmark} & \cellcolor{lightred}\textcolor{black}{\xmark} & \cellcolor{lightgreen}\textcolor{black}{\cmark} & \cellcolor{lightred}\textcolor{black}{\xmark} & \cellcolor{lightred}\textcolor{black}{\xmark} \\
Schopow et al. & 1 & Medicine & Human & \cellcolor{lightgreen}\textcolor{black}{\cmark} & \cellcolor{lightred}\textcolor{black}{\xmark} & \cellcolor{lightgreen}\textcolor{black}{\cmark} & \cellcolor{lightred}\textcolor{black}{\xmark}  & \cellcolor{lightred}\textcolor{black}{\xmark}\\
MedREQAL & 2786 & Medicine & LLM & \cellcolor{lightgreen}\textcolor{black}{\cmark} & \cellcolor{lightgreen}\textcolor{black}{\cmark} & \cellcolor{lightred}\textcolor{black}{\xmark} & \cellcolor{lightgreen}\textcolor{black}{\cmark} & \cellcolor{lightred}\textcolor{black}{\xmark} \\
HealthFC & 750 & Consumer Health & Human & \cellcolor{lightgreen}\textcolor{black}{\cmark} & \cellcolor{lightgreen}\textcolor{black}{\cmark} & \cellcolor{lightred}\textcolor{black}{\xmark} & \cellcolor{lightgreen}\textcolor{black}{\cmark} & \cellcolor{lightred}\textcolor{black}{\xmark} \\
ConflictingQA & 238 & Multi-Domain & LLM & \cellcolor{lightred}\textcolor{black}{\xmark} & \cellcolor{lightred}\textcolor{black}{\xmark} & \cellcolor{lightgreen}\textcolor{black}{\cmark} & \cellcolor{lightred}\textcolor{black}{\xmark} & \cellcolor{lightgreen}\textcolor{black}{\cmark} \\
MedEvidence & \dsetSize{} & Medicine & Human & \cellcolor{lightgreen}\textcolor{black}{\cmark} & \cellcolor{lightgreen}\textcolor{black}{\cmark} & \cellcolor{lightgreen}\textcolor{black}{\cmark} & \cellcolor{lightgreen}\textcolor{black}{\cmark} & \cellcolor{lightgreen}\textcolor{black}{\cmark} \\
\bottomrule
\end{tabular}%
}

\end{table}


An overview of related works and the key distinct contributions of our current work are summarized in Table \ref{tbl:related_work}.

\textbf{LLM-based medical systematic review} 
Numerous studies have explored the potential of LLMs to automate various aspects of scientific literature review, including literature search, query augmentation, screening, data extraction, bias assessment,  narrative synthesis, and answering simple clinical inquiries  \cite{LLMs_for_SR_scopingReview, LLMs_for_synthesis_SR}. %
However, larger-scale evaluations of LLM-based SR or meta-analyses generation remain relatively underexplored. Reason et al. \cite{NMA_caseStudy} examined the ability of LLMs to extract numerical data from abstracts and generate executable code to perform meta-analyses. While their results are promising, the study is limited to just four individual case studies. Schopow et al. \cite{SR_automation_endtoend_caseStudy} and Qureshi et al. \cite{SR_automation_endtoend_caseStudy_webinar} investigate LLM usage across a range of systematic review stages, including meta-review and narrative evidence synthesis, but  also present findings on a very small-case study scale ($ N <10$) and rely on comparison to humans. Overall, these investigations have been limited in scope and require substantial amounts of review from medical experts, highlighting the need for automated benchmarks to help evaluate LLMs' progress. 
 
\textbf{Verification of medical facts derived from systematic reviews} Several studies have leveraged SRs to benchmarked LLMs' ability to perform medical fact verification, where a model must decide whether to support or refute a given claim. For instance, MedREQAL \cite{MedREQAL} is an LLM-curated closed QA dataset designed to investigate how reliably models can verify claims derived from Cochrane SRs. However, it does not provide the sources used by the SRs. Instead,  the dataset evaluates models on their internal knowledge, making the task a form of fact recall. HealthFC \cite{healthfc}, on the other hand, tasks models with verifying claims analyzed by the medical fact-checking site Medizin Transparent, but it only provides pre-synthesized analysis from the web portal as evidence. In contrast to real SR generation, this task primarily involves retrieving information from a pre-synthesized source, removing the real complexity of reasoning across unsynthesized evidence. 
Unlike prior work, MedEvidence requires extracting, reasoning over, and synthesizing relevant information across single or multiple sources (each with different levels of evidence) to match the expert-derived conclusion of a SR (without access to the original SR itself). It resembles the intricacies of SR analysis, as the raw sources (articles/abstracts) are directly provided to the model.


\textbf{LLM Behavior in the Presence of Conflicting Sources} 
ConflictingQA \cite{conflictingQA} examines how models respond to conflicting arguments supporting or refuting a claim. However, it focuses on inherently contentious questions without definitive answers, spans domains beyond medicine, and uses diverse online sources rather than peer-reviewed literature. ClashEval \cite{clasheval} investigates conflicts between a model's internal knowledge and external evidence, including a drug-related (medical) subset, but limits evaluation to single-source conflicts with artificially perturbed values. ConflictBank \cite{conflictBank} and KNOT \cite{knot_conflict} assess model performance on specific conflict types—such as temporal inconsistencies, misinformation, and logic-based contradictions—but rely on factoid-style questions sourced from Wikipedia. These benchmarks only leverage relatively small and synthesized inputs.

To the best of our knowledge, no existing studies or datasets provide richly annotated data to systematically benchmark models' ability to align with the conclusions of medical systematic reviews while using the same underlying research documents as the original medical experts.

\section{Dataset Curation Process}

\textbf{Data provenance} We collect open-source systematic reviews, available via PubMed, conducted by Cochrane, an international non-profit organization dedicated to synthesizing evidence on healthcare interventions through contributions from over 30,000 volunteer clinician authors \cite{how_to_cochrane}. Cochrane is a long-standing and widely respected source of clinical evidence \cite{cochrane_quality, what_is_cochrane}, offering open-access content and analyses presented in a standardized format. Additionally, for each SR, we collect all the cited studies that are relevant for a given conclusion (we refer to these studies as `sources'). When the source article's full text is available (i.e. the article is open-source), we obtain it using the existing BIOMEDICA dataset \cite{biomedica}; otherwise, abstracts are retrieved directly via PubMed's Entrez API \cite{entrez}. All retrieved full-text articles use a CC-BY 4.0 license, which allows for re-distribution.

\begin{figure}
    \centering
    \includegraphics[width=1.0\linewidth]{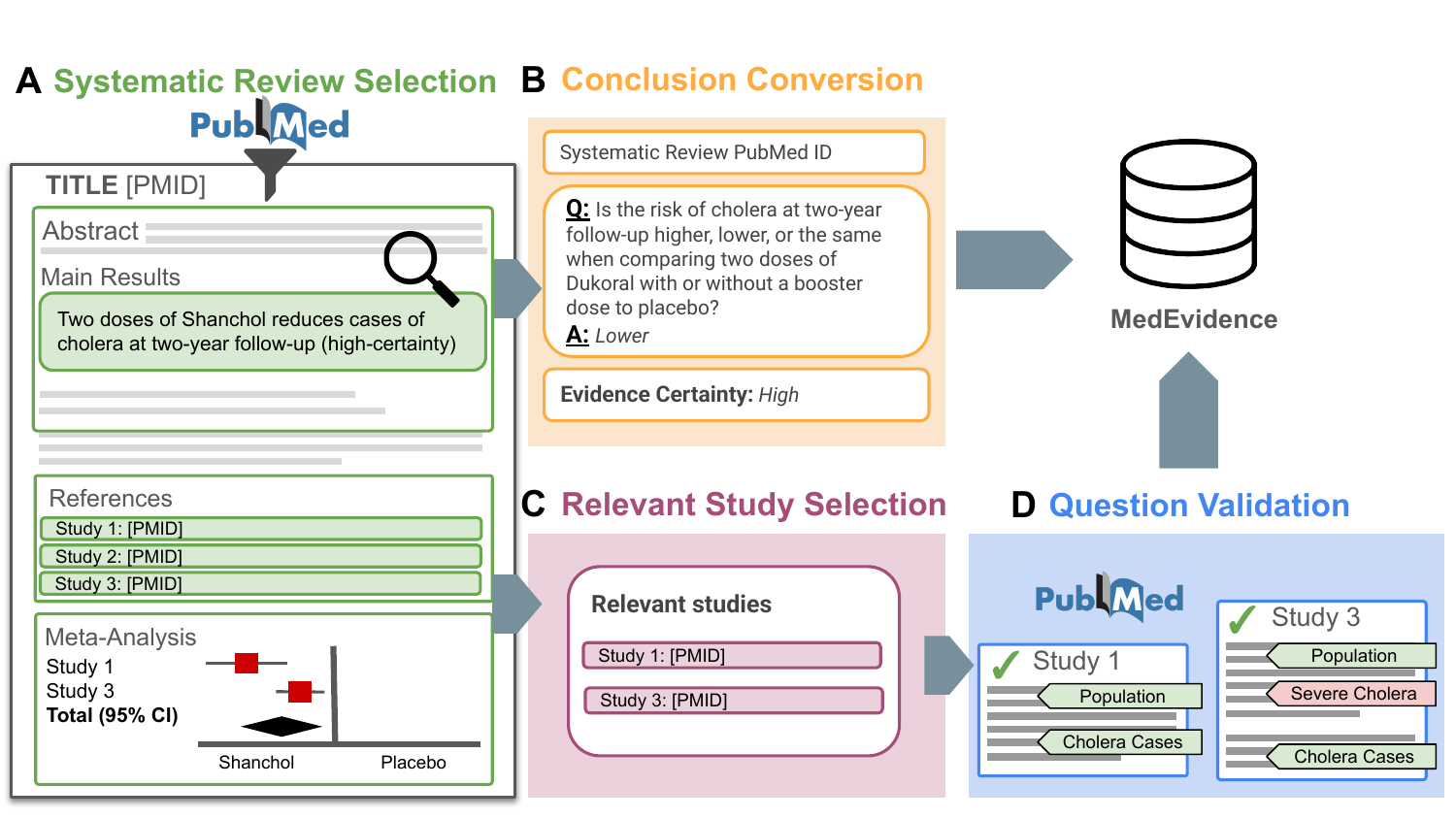}
    \caption{Overview of the dataset curation process for \dsetName{}.}
    \label{fig:data_collection}
\end{figure}

\textbf{Dataset curation pipeline} The core challenge in creating our dataset is ensuring that an LLM is provided with sufficient information to reproduce a given conclusion. To ensure a high-quality dataset, we developed a four-stage pipeline consisting of: (1) systematic review selection, (2) conclusion to questions conversion, (3) relevant study selection, and (4) question feasibility validation (as shown in Figure \ref{fig:data_collection}).


\begin{enumerate}
    \item  \textbf{Systematic review selection} We use Entrez to retrieve all Cochrane SRs published between January 1, 2014 to April 4, 2024 \cite{pubmed_sb}. We only include  systematic reviews for which all sourced studies are indexed in PubMed (with at least an abstract available). We additionally retrieve all data and metadata for the sourced studies, including: full-text via BIOMEDICA (when it is available), abstract, mesh terms, title, and publish date. 

    \item \textbf{Conclusion to question conversion.} Cochrane reviews follow a standardized format, allowing for a systematic conversion process. To identify potential questions, we followed the protocol below: Human annotators were instructed to review the SR abstract and examine the "Main Results" subsection (see Appendix Figure \ref{fig:ex-mainResultsSection} for an example) to identify individual conclusive statements that statistically compare an intervention with a control group. These individual statements were then converted into question–answer pairs by the annotators, with answers belonging to a fixed set of classes. To be clear, \texttt{insufficient data} was used for statements by the SR authors explicitly indicating that no study investigated—or included sufficient data to analyze—the combination of treatment, control, and outcome; \texttt{uncertain effect} referred to cases where analysis was performed but definitive conclusions could not be made (see Appendix Section \ref{a:c2a_conversion} for more conversion details). Evidence certainty was extracted only when it was explicitly provided by the original SR authors, who use the standardized GRADE framework \cite{bezerra2022assessment} to assess the quality of evidence in the included studies. This certainty is often stated in the abstract, indicating the strength or quality of each observation. 
    
    \item  \textbf{Relevant study selection} To identify relevant studies for a given SR, annotators used the analysis section provided in the appendix, which "weighs" the contributions of sources supporting each conclusion. For questions with insufficient data (where it is not possible to determine weights), reviewers were instructed to include studies cited in the SR that either (1) discuss the specified treatment and control but not the outcome, or (2) evaluate the treatment and outcome but compare against a different control.
    
    \item  \textbf{Question feasibility validation} Finally, given the question–answer pair and the source studies, annotators were tasked with determining whether the question was answerable based on the provided information. A question was considered answerable if at least 75\% of the total weight in the analysis came from "valid" studies included in the meta-analysis. We define a study as "valid" if it (1) provides numerical data on both the intervention and control groups specified in the question, and (2) includes statistical or numerical details about the difference between the groups on the specified outcome—such as raw counts, p-values, confidence intervals, or risk ratios. The most common reason for discarding conclusions was when review authors pooled outcome data across studies, but the outcome was omitted or discussed without clear statistical detail in the abstracts of relevant studies.
\end{enumerate}

 In addition to these human-curated metadata, we use an LLMs to assess the percentage of individual source studies whose answer to the question aligns with the final answer provided in the systematic review. Thus, to calculate source-level agreement (which we call `source concordance') we prompt DeepSeekV3 (the strongest model in our benchmark) to answer the question using only one single relevant source; the source is deemed to `agree' with the final answer if and only if the LLM's classification with the one source matches the ground truth classification. 

\textbf{Medical domain taxonomy assignment} To identify the relevant medical specialties in our dataset, we extract the Medical Subject Headings (MeSH terms)—a controlled vocabulary used by PubMed to index papers—from the \dsetReviews{} systematic reviews included in our dataset. We then feed this list into DeepSeek to generate a simplified categorization of specialties, resulting in \nSpecialties{} categories. Finally, we prompt DeepSeek to assign each question to the most relevant category, or to an "Other" category if no specific specialization is applicable.
\section{Dataset Description}
\label{sec: Dataset-Description}

\begin{figure}
    \makebox[\textwidth][c]{\includegraphics[width=1.3\textwidth]{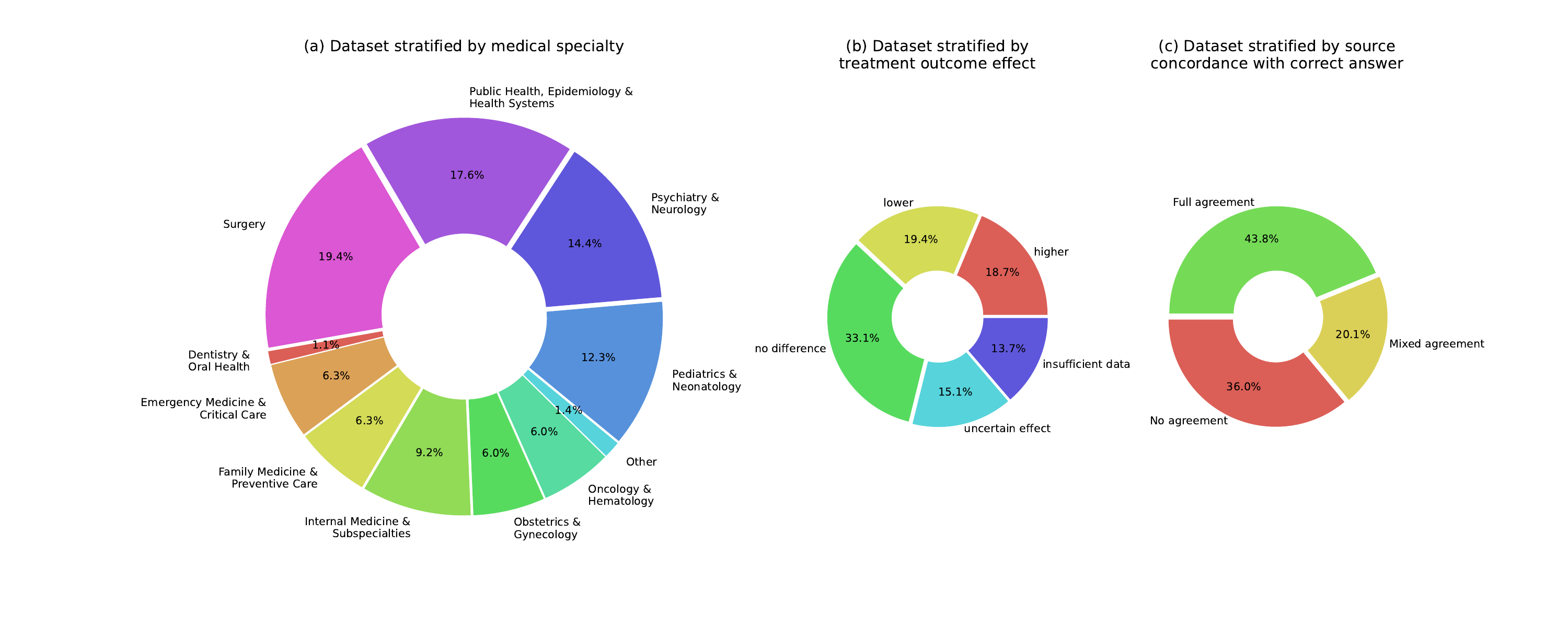}}
    \caption{Key statistical characteristics of the questions in \dsetName{}. (a) shows the dataset distribution stratified by medical specialty. (b) presents the distribution stratified by outcome effect. (c) shows the distribution stratified by source concordance with the expert-assessed treatment outcome effect (i.e. the correct answer).}
    \label{fig:data-distrib}
\end{figure}
\begin{table}[h]\small
\caption{Sample question from the dataset. Fields marked with an asterisk (*) use LLMs to assist the generation. Relevant source details are omitted here for brevity.}
\setlength{\tabcolsep}{0.9mm}{
\begin{tabular}{p{0.35\linewidth} | p{0.64\linewidth}}
\toprule
\textbf{Question}   & Is stroke prevention higher, lower, or the same when comparing Transcatheter Device Closure (TDC) to medical therapy? \\
\textbf{Answer}                        & no difference \\
\textbf{Relevant Sources (PubMed IDs)} & 22417252, 23514285, 23514286 \\
\textbf{Systematic Review (PubMed ID)} & 26346232  \\
\textbf{Review Publication Year}  & 2015 \\
\textbf{Evidence Certainty}              & n/a  \\
\textbf{Open-Access Full-Text Needed}  & no   \\
\textbf{*Source Concordance}  & 1.0           \\
\textbf{*Medical Specialty}  & Surgery           \\
\bottomrule
        \end{tabular}}
\label{tbl:sample_q}
\end{table}

\dsetName{} contains a total of \dsetSize{} questions derived from \dsetReviews{} systematic reviews with \dsetIncludedRefs{} referenced individual articles, of which \dsetFulltextRefs{} have full-text available (see Appendix Figure \ref{fig:data-filtering} for a cohort diagram of the dataset). Questions were systematically collected by three human annotators with between one and five years of graduate education. Figure \ref{fig:data-distrib} shows the dataset distribution stratified by specialty, outcome effect, and source concordance with the expert-assessed treatment outcome effect (i.e. the correct answer). The benchmark covers topics from  \nSpecialties{} medical specialties (e.g. public health, surgery, family medicine, etc.), five different outcome effects (\texttt{higher}, \texttt{lower}, \texttt{no difference}, \texttt{uncertain effect}, \texttt{insufficient data}), and three broad levels of concordance between the source paper and the correct answer (full agreement, no agreement, mixed agreement). Additional characteristic distributions of the dataset can be found in Appendix Figure \ref{fig:extra-data-distrib}.

\textbf{Data format.} \dsetName{} is grouped by question; each question includes core data for evaluation, metadata, as well as the content details for the relevant sources. The core data consists of: a human-generated question of the form ``Is [quantity of medical outcome] higher, lower, or the same when comparing [intervention] to [control]?"; the taxonomized answer to the question (\texttt{higher}, \texttt{lower}, \texttt{no difference}, \texttt{uncertain effect}, \texttt{insufficient data}); and the list of relevant studies (sources) used by the review authors to perform the analysis, identified by their unique PubMed IDs. We additionally provide the following metadata: the systematic review from which the question was extracted; the publication year of the systematic review; the authors' confidence in their analysis, also referred to as the `evidence certainty' (\texttt{high}, \texttt{moderate}, \texttt{low}, \texttt{very low}, or \texttt{n/a} if not provided); a Boolean identification of whether full-text is available and needed to answer the question; the exact fractional source concordance; and the medical specialty associated with the question. Separately, for each source, we provide the unique PubMed ID, title, publication date if available, and content (full-text if available in PMC-OA, abstract otherwise). An individual data point example is shown in Table \ref{tbl:sample_q}.

\begin{figure}
    \centering
    \includegraphics[width=1\linewidth]{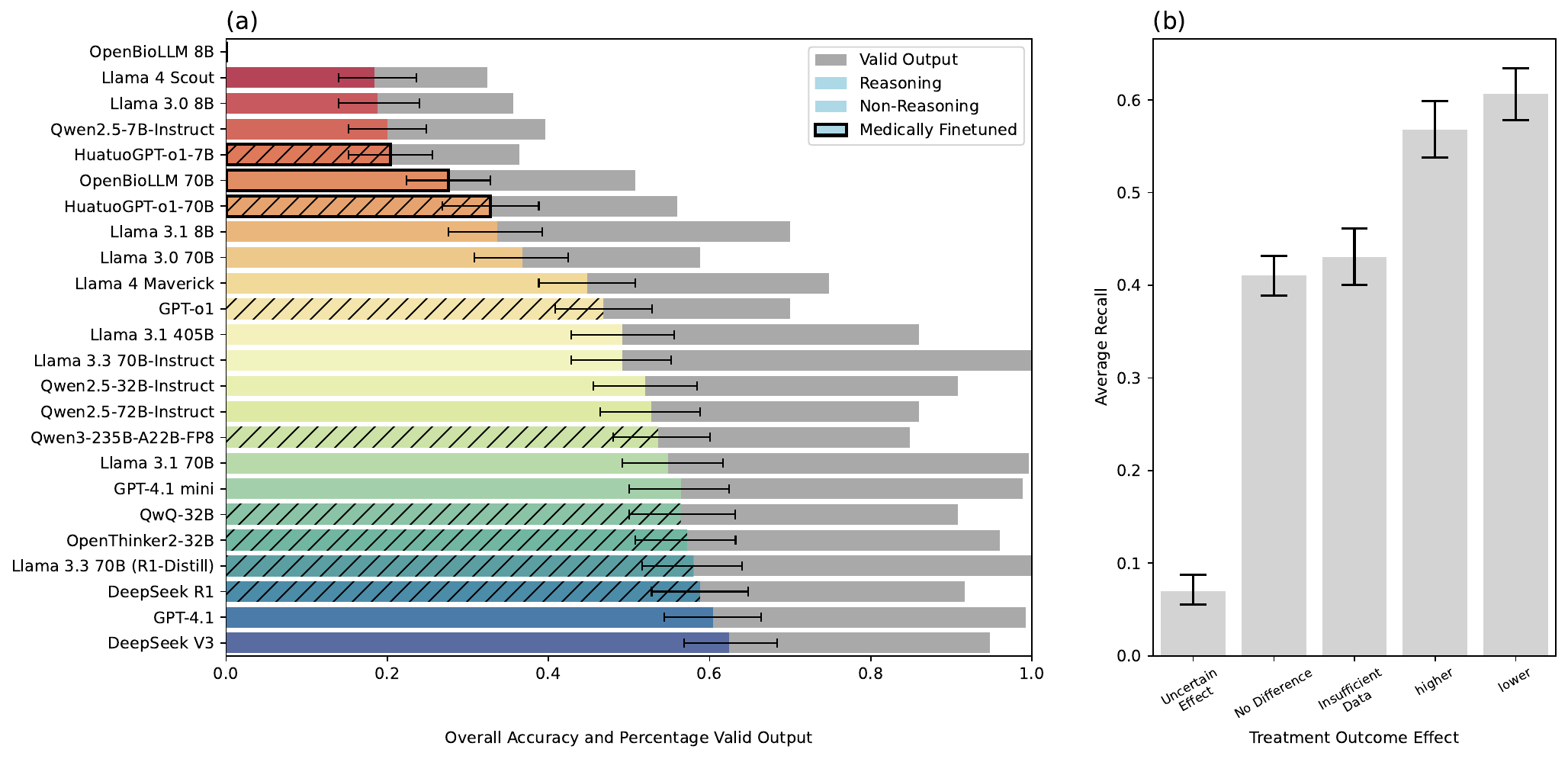}
    \caption{(a) Average model accuracy (and 95\% CI) on \dsetName{}, overlaid on the percentage of questions where the model provided valid output (additional details in Appendix Section \ref{a:sec:instr-follow}). (b) Average recall by ground truth treatment outcome effect, aggregated across all models (with overall 95\% interval). Per-model average recall by treatment outcome effect can be found in Appendix Figure \ref{fig:per-class-recall}.}
    \label{fig:base-acc_avg-perclass-recall}
\end{figure}


\section{Benchmarking LLM performance}
\label{sec:Benchmarking-LLM-Performance}

\subsection{Experimental settings}
\label{sec: Experimental settings}
\textbf{LLM selection} We selected \modelCount{} LLMs across different configurations, including a variety of sizes (from 7B to 671B), reasoning and non-reasoning capabilities, commercial and non-commercial licensing, and medical fine-tuning. This selection includes GPT-o1 \cite{gpto1}, DeepSeek R1 \cite{deepseekR1},  OpenThinker2 \cite{openthoughts}, GPT-4.1 \cite{gpt4}, Qwen3 \cite{qwen3}, Llama 4 \cite{llama4}, HuatuoGPT-o1 \cite{huatuogpt-o1},  OpenBioLLM \cite{OpenBioLLMs}, and more (please see Appendix Table \ref{tbl:model-info} to see details of all selected models). This selection is non-exhaustive; rather, it is designed to investigate overarching trends across different model types.


\textbf{Prompting setup} 
\begin{enumerate}
    \item  \textbf{Basic prompt} We evaluated all models in a zero-shot setting, prompting them to first provide a rationale for their answer, followed by an `answer' field containing only one option from the list of five valid treatment outcome effects (higher, lower, no difference, uncertain effect, or insufficient data). To assess the models’ ``natural'' behavior, we provided minimal guidance in the prompt beyond specifying the required response format, and supplied the abstracts or full text of the relevant studies as context (see Appendix Figure \ref{fig:basic-prompt}).

\item \textbf{Expert-guided prompt} LLMs may not natively understand how to handle multiple levels of evidence, which can lead to unfair evaluations. To address this, we explicitly design a prompt that instructs the LLM to summarize the study design and study population, and to assign a grade of evidence based on established definitions of grades of recommendation (see Appendix Figure \ref{fig:expert-prompt} for the full prompt).

\end{enumerate}

\begin{figure}
    \makebox[\textwidth][c]{\includegraphics[width=1.2\textwidth]{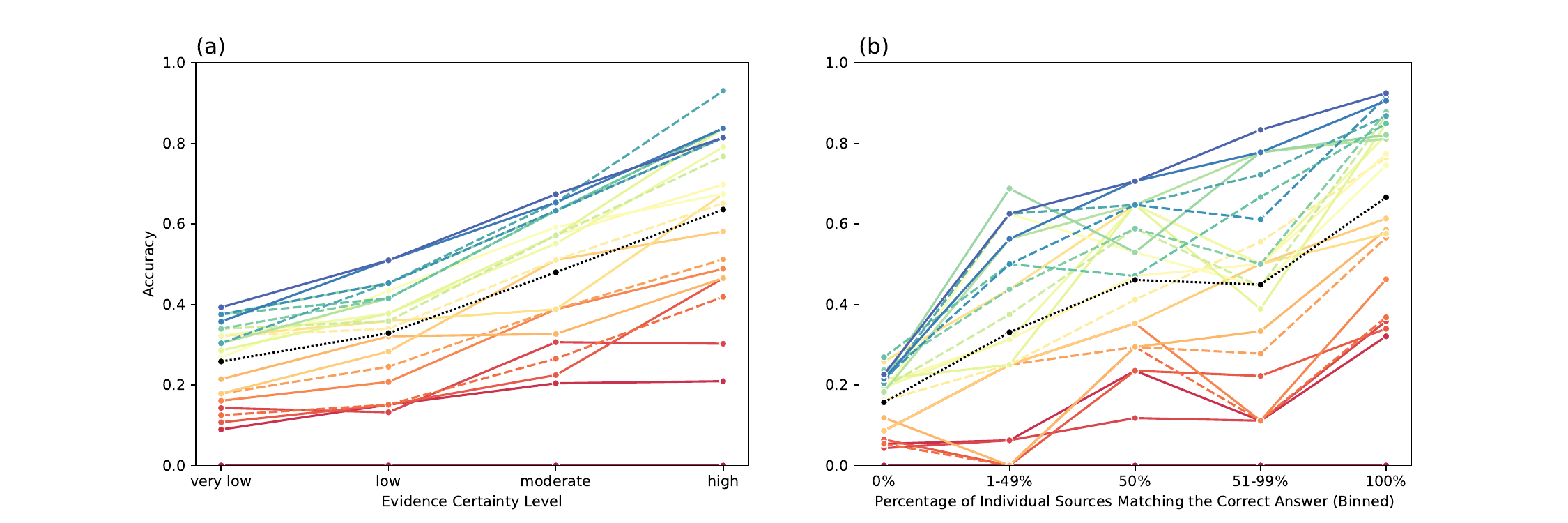}}
    \caption{(a) Accuracy as a function of evidence certainty, shows a monotonically increasing trend. (b) Accuracy as a function of source concordance, defined as the percentage of relevant sources that agree with the final systematic review (SR) answer, also exhibits a monotonically increasing trend.}
    \label{fig:certainty-sla}
\end{figure}



For both cases, if the input exceeded the LLM’s context window, we used multi-step refinement (via LangChain's RefineDocumentsChain \cite{langchain_refine_documents_chain}) to iteratively refine the answer based on a sequence of article chunks. All models were evaluated with zero temperature to maximize reproducibility. 

\textbf{LLM evaluation} Model performance was evaluated using accuracy based on an exact match between the answer field and the ground truth. Model outputs were lower-cased and stripped of whitespace before comparison. If no `answer' field was provided, or if its content was not an exact rule-based match with the correct answer, the output was deemed incorrect. Confidence intervals (CIs) were calculated via bootstrap (95\%, N=1000) \cite{efron1994introduction}. 

\textbf{Compute Environment} Experiments were performed in a local on-prem university compute environment using 24 Intel Xeon 2.70GHz CPU cores, 8 Nvidia H200 GPUs, 16 Nvidia A6000 GPUs, and 40 TB of Storage. Large-scale models that could not be run locally in this environment were queried in the cloud using public APIs available from together.ai or OpenAI. 

\label{sec:baseline-performance}


\begin{wrapfigure}{r}{0.45\linewidth}
    \centering
     \includegraphics[width=\linewidth, trim=0.8em 0.7em 0.6em 0.3em, clip]{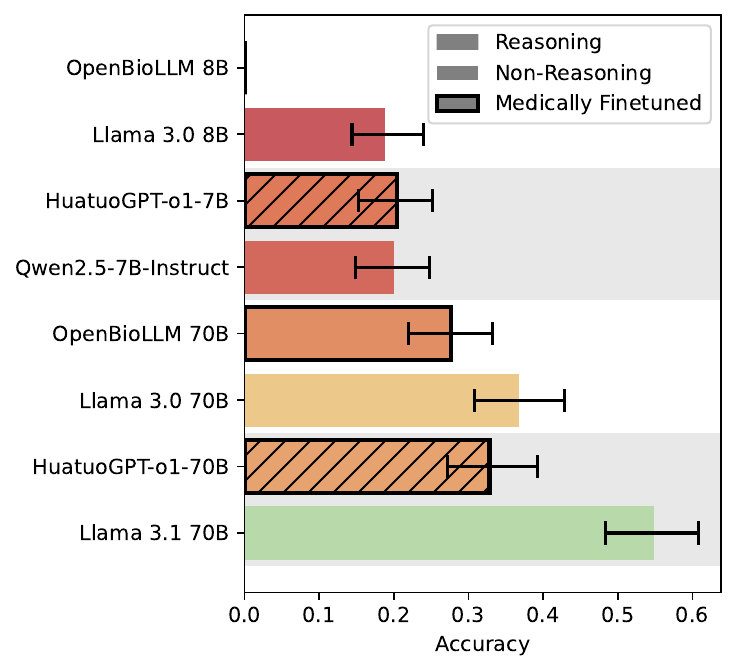}
    \caption{Medically-finetuned models vs their base generalist counterparts. Pairs of medical and base models are adjacent. 95\% confidence intervals are calculated via bootstrapping with $N=1000$.}
    \label{fig:med-vs-base}
\end{wrapfigure}


\section{Discussion}
As shown in Figure \ref{fig:base-acc_avg-perclass-recall} (a), even frontier models such as DeepSeek V3 and GPT-4.1 demonstrate relatively low average accuracy of 62.40\% (56.35, 68.45) and 60.40\% (54.30, 66.50), respectively—far from saturating our benchmark.  We identify four key factors that influence model performance on our benchmark: (1) token length, (2) dependency on treatment outcomes, (3) inability to assess the quality of evidence, and (4) lack of skepticism toward low-quality findings. Additionally, we found that (5) medical fine-tuning does not improve performance, and (6) model size shows diminishing returns beyond 70 billion parameters. We explore each of these factors in more detail below using the basic prompt setup.

\textbf{Reasoning vs non-reasoning LLMs}
We highlight that, in general, reasoning models do not consistently outperform non-reasoning models of the same class or size on MedEvidence (Figure  \ref{fig:base-acc_avg-perclass-recall} (a)), as evidenced by DeepSeek V3 outperforming its reasoning counterpart (DeepSeek R1), while LLaMA 3.3 70B distilled from DeepSeek R1 outperforms the LLaMA 3.3 70B base model.

\textbf{Model performance decreases as token length increases}
Generally, performance on MedEvidence drastically reduces as the number of tokens increases (Appendix Figure \ref{fig:num-tokens}).  Naturally, training LLMs on long contexts does not guarantee improved long-context understanding, as models may still struggle to utilize information from lengthy inputs \cite{chen2024long,li2404long}. 


\textbf{Model performance dependency on treatment outcome effect}
Figure \ref{fig:base-acc_avg-perclass-recall} (b) shows the per-class recall stratified by treatment outcome effect. Overall, all models perform best on questions where the correct answer corresponds to \texttt{higher} or \texttt{lower} effects—cases where a strong stance can be taken. They are slightly less successful on \texttt{no difference} and \texttt{insufficient data} questions, where a definitive conclusion is available but there is no clear preference for either treatment. Performance is lowest on the most ambiguous class, \texttt{uncertain effect}. Notably, as shown in Appendix Figure \ref{fig:avg-confusion-matrix}, models are generally reluctant to express uncertainty, often committing to a more certain outcome that appears plausible. Notably, previous work has observed LLMs are verbally overconfident  \cite{sun2025large,xiong2023can} and shown that reinforcement learning via human feedback (RLHF) amplifies this effect \cite{leng2024taming}.

\textbf{Model performance improves with increasing levels of evidence} We leverage the evidence certainty levels reported by experts in each systematic review (SR). As shown in Figure \ref{fig:certainty-sla}(a), the overall ability of models to match SR conclusions improves as the level of evidence increases. We therefore explore whether model performance is also associated with the level of source concordance. As shown in Figure \ref{fig:certainty-sla}(b), models’ ability to match human conclusions increases as the proportion of sources agreeing with the correct answer increases (e.g., DeepSeek V3 achieves 92.45\% accuracy at 100\% source agreement vs. 41.21\% at 0\% source agreement). This suggests that, unlike human experts, current LLMs struggle to critically evaluate the quality of evidence and to remain skeptical of results.
We observe that this behavior persists even when models are prompted (using the expert-guided prompt) to consider study design, population, and level of evidence (Appendix Figure \ref{fig:expert-certainty}).


\textbf{Medical finetuning does not improve performance} Figure~\ref{fig:med-vs-base} compares the average performance of medically finetuned models to their base model counterparts. Across all comparisons, medical finetuning fails to improve performance (even for medical-reasoning models) and, in most cases, actually degrades it.  
Indeed, fine-tuning without proper calibration can harm generalization, sometimes resulting in worse performance than the base model \cite{mai2024fine, kong2020calibrated, wu2024finetunebench}. Similar behavior has been previously reported in long-context medical applications \cite{fleming2024medalign}.


\begin{figure}  
    \centering
    \includegraphics[width=0.9\linewidth]{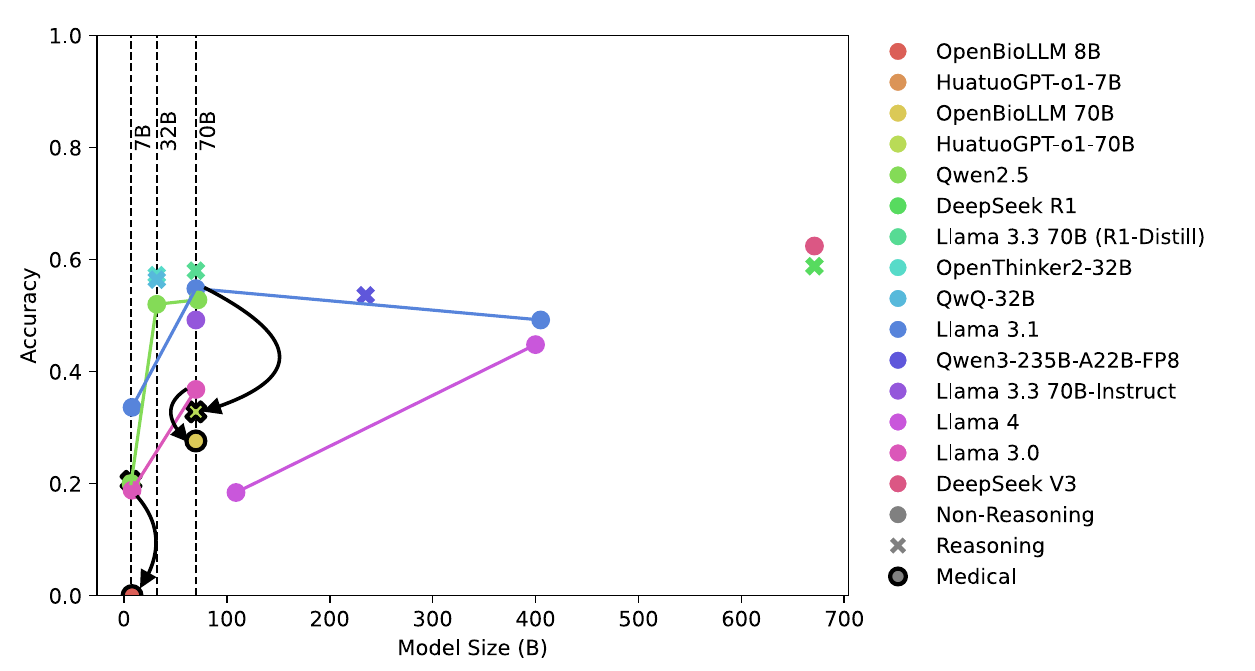}
    \caption{ Average model accuracy as a function of model size. We observe diminishing returns beyond 70 billion parameters. Arrows point from base models to their medically-finetuned counterparts (arrow between HuatuoGPT-o1 7B and Qwen2.5 7B omitted due to very similar performance).}
    \label{fig:model-size}
\end{figure}

\textbf{Model size shows diminishing returns beyond 70B parameters} 
As shown in Figure~\ref{fig:model-size}, within the same model families, increasing size from 7B to 70B parameters yields  substantial accuracy gains on MedEvidence. However, beyond this point, we observe rapidly diminishing returns, both within specific model families and across our suite of evaluated models more broadly.

Combined, our results suggest that synthesizing information across sources to match individual systematic reviews' conclusions eludes current scaling paradigms. Increasing test-time compute (i.e., reasoning) does not necessarily improve performance, larger models do not consistently yield greater gains, and knowledge-based fine-tuning tends to degrade performance. Instead, most models exhibit similar behavior:  model performance tends to degrade as token length increases, their responses show overconfidence, and all models exhibit a lack of scientific skepticism toward low-quality findings.
These results suggest that more work is still required before LLMs can reliably match the observations from expert-conducted SRs, even though LLM systems are already deployed and being used by clinicians.


\textbf{Limitations}
Our study has several limitations. First, the dataset is subject to selection bias, as we only include a SR if all its sources are available (either full text/abstract). Second, while our benchmark is designed to isolate and provide a controlled environment to test LLMs' ability to reason over the same studies experts used to derive  conclusions, it does not assess the full SR pipeline, including literature search, screening, or risk-of-bias assessment.  Future work could incorporate multi-expert consensus or update findings based on newer studies to strengthen benchmark reliability.


\section{Conclusion}
Benchmarks drive advancements by providing a standard to measure progress and enabling researchers to identify weaknesses in current approaches. While LLMs are already deployed for scientific synthesis, our understanding of their failure modes still requires broader investigation. In this work, we present MedEvidence, a benchmark derived from gold-standard medical systematic reviews. We use MedEvidence to characterize the performance of 24 LLMs and find that, unlike humans, LLMs struggle with uncertain evidence and cannot exhibit skepticism when studies present design flaws. Consequently, given the same studies, frontier LLMs fail to match the conclusions of systematic reviews in at least ~37\% of evaluated cases.  We release MedEvidence to enable researchers to track progress.

\newpage
\bibliographystyle{unsrt}
\bibliography{main}

\newpage

\newpage
\appendix

\section*{Acknowledgements}
We thank Dr. Jeffrey Nirschl and Professor Robert Tibshirani for their invaluable discussions...

\section*{Appendix contents}
\begin{adjustwidth}{2.5em}{0pt}
    \startcontents[sections]
    \printcontents[sections]{l}{1}{\setcounter{tocdepth}{2}}
\end{adjustwidth}

\section{Societal impact}

The use of large language models  to automate systematic reviews offers clear potential to accelerate evidence synthesis in medicine and policy. However, when these systems produce incorrect or misleading results, clinicians and policymakers may base decisions on flawed findings, leading to inappropriate treatments or misguided recommendations.

Our study underscores the urgent need for continued research and cautious deployment. LLM-based  systematic review systems need further rigorous validation, transparent uncertainty quantification, and mechanisms to detect and mitigate biases and errors. Only through careful development and oversight can these technologies be harnessed to benefit society without exacerbating existing risks or creating new harms.

\section{Dataset collection details}
Below, we provide additional in-depth details regarding stages in dataset curation process.

\subsection{Systematic review selection}
\begin{figure}[hbt]
    \centering
    \includegraphics[width=0.4\linewidth]{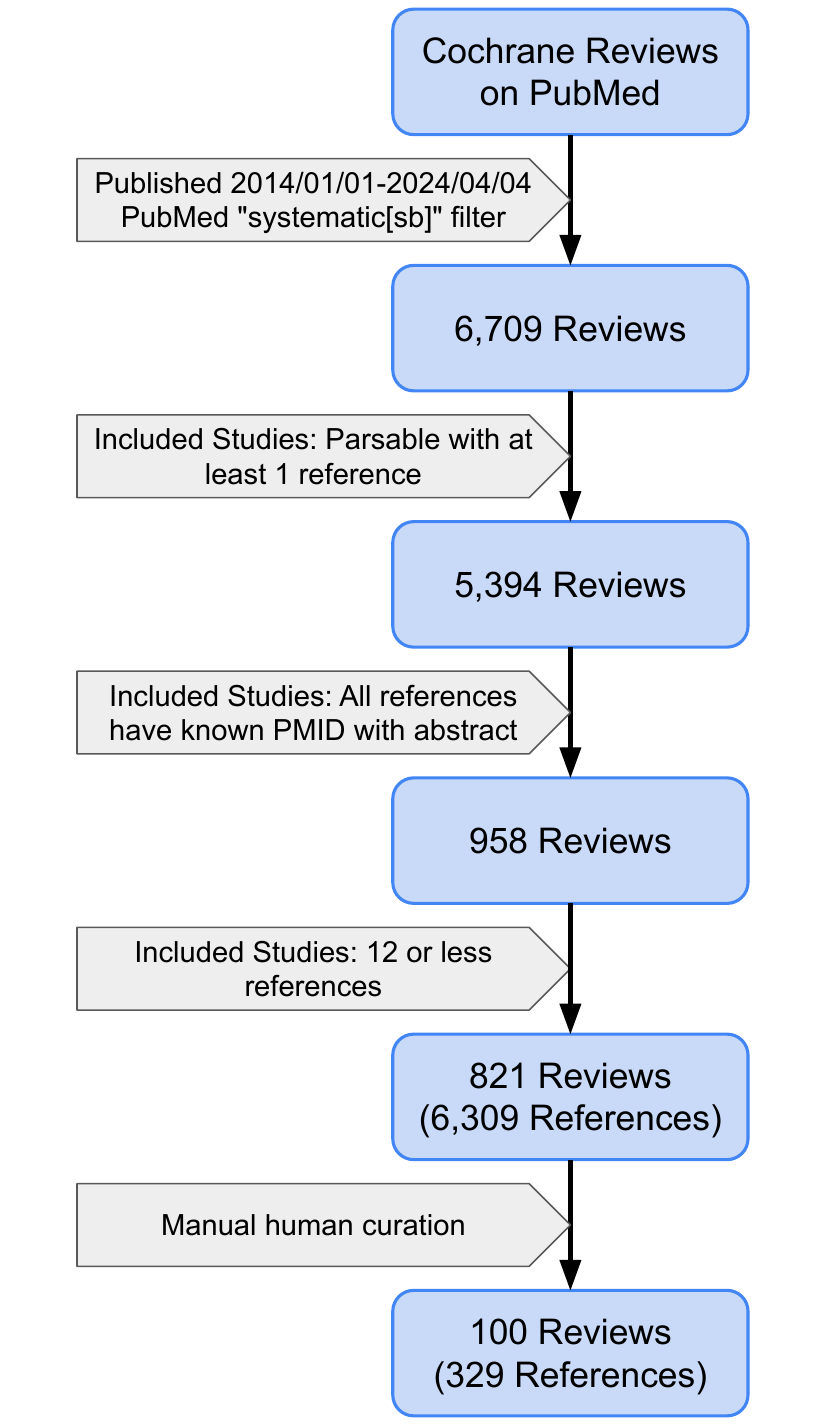}
    \caption{\dsetName{} cohort diagram describing selection criteria for Cochrane SRs suitable for use in the \dsetName{} dataset. Note that not all available papers in the second-to-last stage were manually reviewed for use in the final stage.}
    \label{fig:data-filtering}
\end{figure}

\dsetName{} is originally derived from 6,709 Cochrane publications extracted via Entrez from PubMed. We first discarded any papers where first References subsection was not both entitled ``Studies included in this review" and non-empty, as our initial extraction filter included Cochrane SR protocols and SRs finding no valid studies, which were not of interest. We filter for SRs where all included references have a retrievable abstract and limit to SRs with 12 or less references to reduce annotator burden and improve odds of finding SRs where questions can be validated. On average, the end-to-end creation of a single question requires approximately 20 minutes. Appendix Figure \ref{fig:data-filtering} presents a cohort diagram for the materialization of the dataset.

\subsection{Conclusion to question conversion}
\label{a:c2a_conversion}
\begin{figure}
    \centering
    \includegraphics[width=1\linewidth]{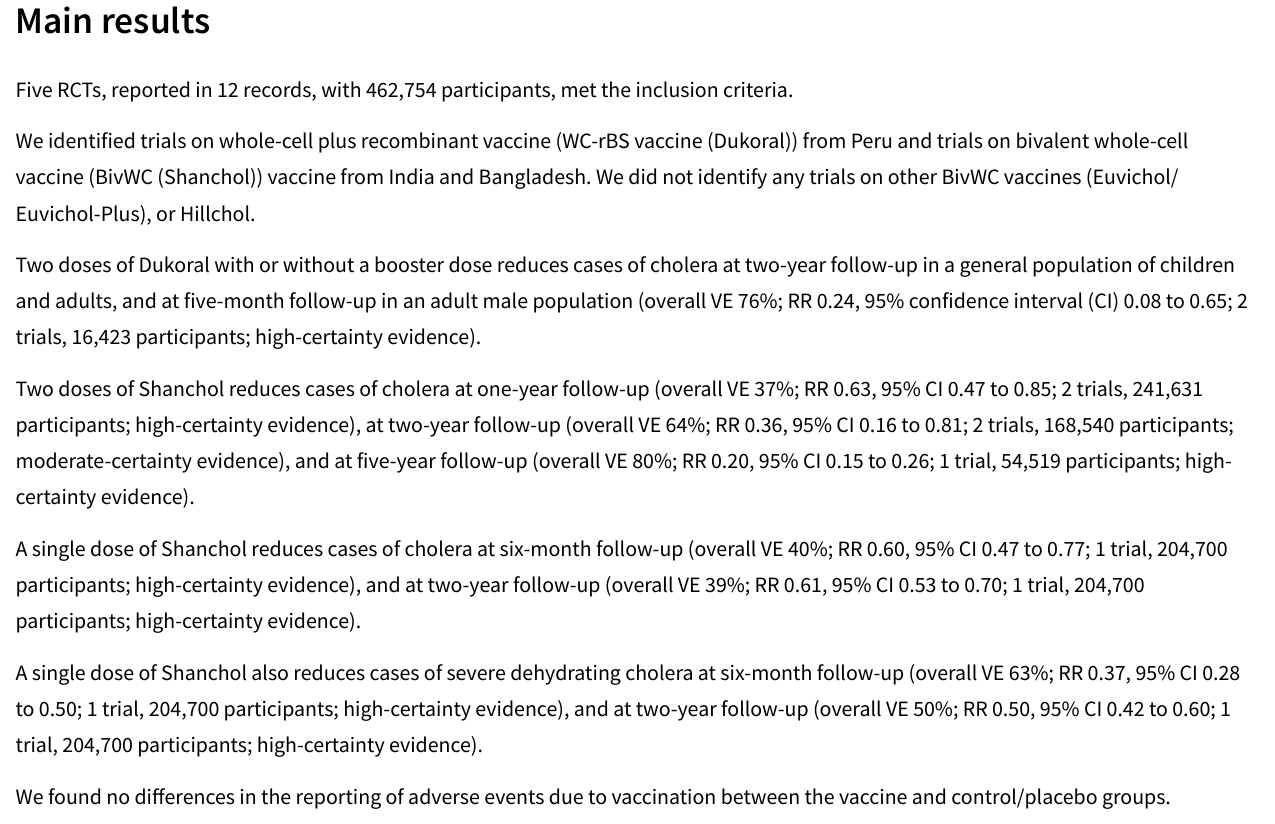}
    \caption{An example of a ``Main Results" section from a Cochrane review used in \dsetName{} (DOI: https://doi.org/10.1002/14651858.CD014573). Annotators were instructed to extract conclusions from this standardized sub-section of the SR abstract.}
    \label{fig:ex-mainResultsSection}
\end{figure}
Appendix Figure \ref{fig:ex-mainResultsSection} provides a direct example of a SR abstract parsed for manual question creation. We highlight the explicit statements (`conclusions') asserting differences between a treatment and control on an outcome, and the presence of standardized, author-provided assessment of evidence certainty for these individual conclusions. SR abstracts were consistently written in this form, allowing annotators to consistently interpret the conclusion into a question. To define the correct answer to the generated question, annotators obeyed the following criteria:
\begin{itemize}
    \item Outcomes, or pairs of treatments and controls, where the authors stated that no studies provided sufficient (or any) evidence to perform analysis were labeled as \texttt{insufficient data} questions.
    \item Conclusions in which the authors stated that there was ``no difference'' or ``no significant difference'' between treatments and controls were labeled as \texttt{no difference} questions.
    \item Conclusions where the authors stated a difference between outcomes either definitively or with qualification (e.g. `X increases Y' or `X may reduce Y') were given the appropriate \texttt{higher} or \texttt{lower} label.
    \item Conclusions where the authors expressed that uncertainty was too great to evaluate a treatment outcome effect were placed in the \texttt{uncertain effect} label class. Conclusions where authors assessed a difference, but then stated that they were very uncertain of their findings were deemed ambiguous and discarded.
\end{itemize}

\subsection{Relevant study selection and question validation}
For author conclusions where more than one study was used, SRs provide meta-analyses over all relevant sources (an example meta-analysis is shown in Appendix Figure \ref{fig:ex-metaAnalysisSection}), allowing us to confirm whether the studies used in the original SR contain sufficient information to replicate the conclusions of human analysis.
\begin{figure}
    \centering
    \includegraphics[width=1\linewidth]{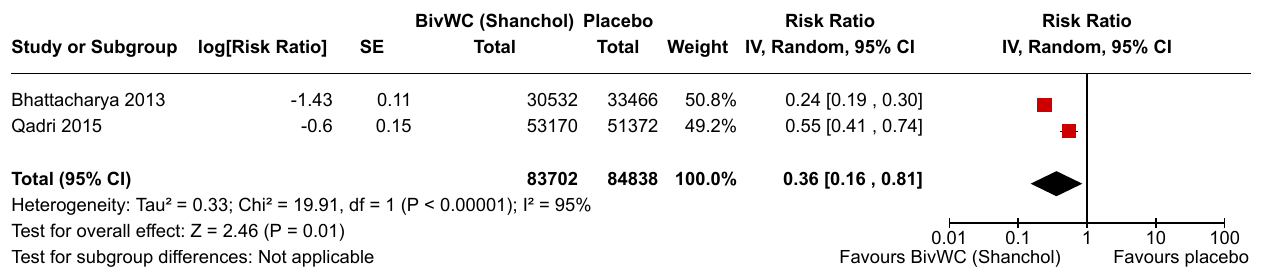}
    \caption{An example meta-analysis from a Cochrane review (figure from DOI: https://doi.org/10.1002/14651858.CD014573). Notably, the set of relevant studies and their individual weighted contributions to the overall result are available.}
    \label{fig:ex-metaAnalysisSection}
\end{figure}


\FloatBarrier

\section{Additional dataset distributions}

\begin{figure}
    \centering
    \includegraphics[width=1\linewidth]{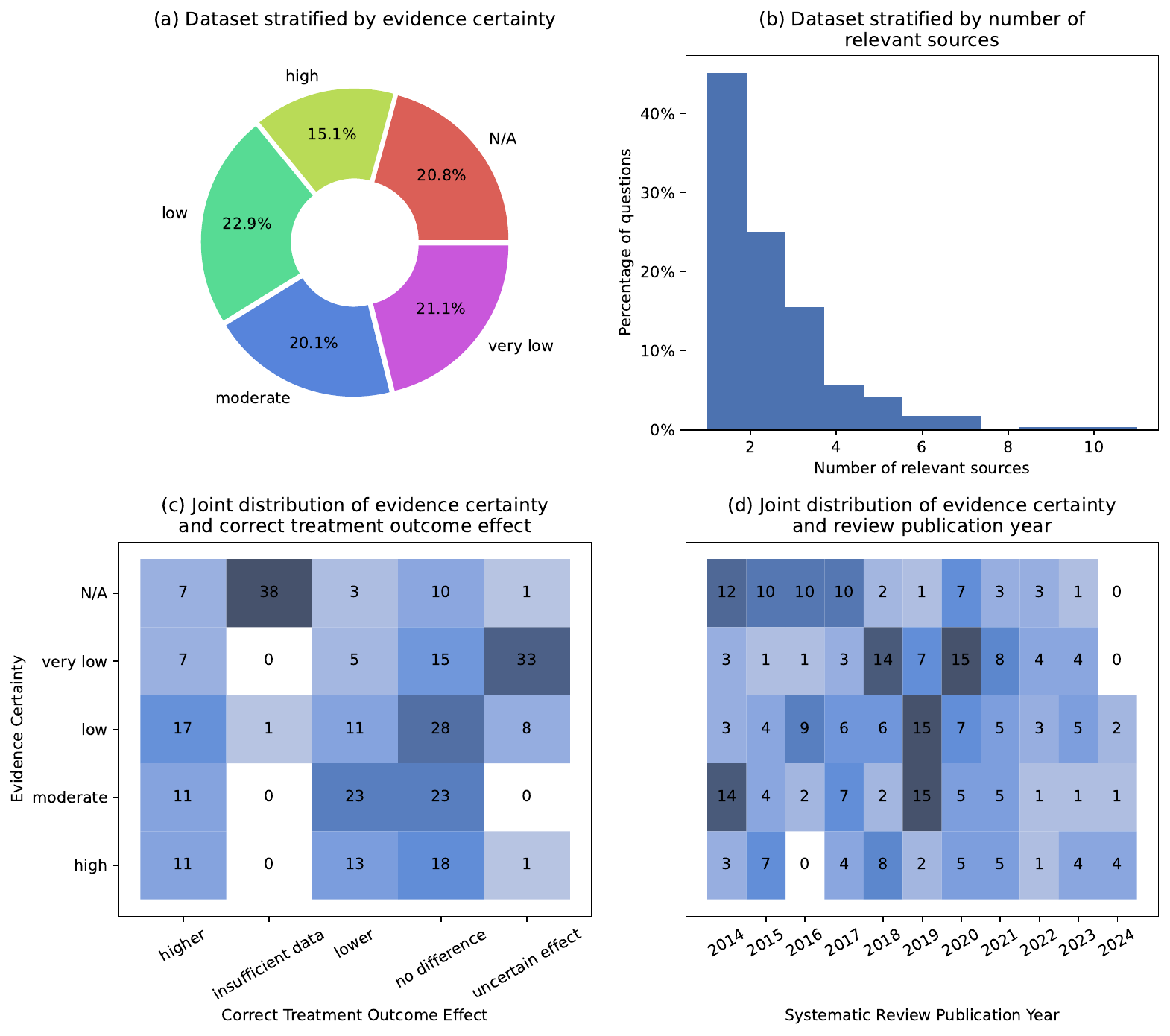}
    \caption{Additional statistical characteristics of \dsetName{}. (a) shows the dataset distribution stratified by evidence certainty. (b) stratifies the questions by number of relevant sources. (c) is a joint distribution of evidence certainty and correct answer label. (d) shows the distribution of evidence certainties by systematic review publication year.}
    \label{fig:extra-data-distrib}
\end{figure}
We present additional statistical characteristics of the questions in our \dsetName{} dataset in Appendix Figure~\ref{fig:extra-data-distrib}. We highlight that the dataset is balanced with respect to evidence certainty levels, strengthening the reliability of our main observations on the relationship between evidence certainty and model performance. With regard to the joint distribution of correct treatment outcome effect and evidence certainty, we note that the highly concentrated distributions for the \texttt{insufficient data} and \texttt{uncertain effect} classes are inherent to the nature of SR. For example, in the case of the \texttt{insufficient data} class, authors cannot draw definitive conclusions from analyses they were unable to perform; thus, their findings are most uncertain when the quality of evidence is poor.

\section{Evaluated models and prompts}
\label{a:sec:models_prompts}
The full list of \modelCount{} models we evaluate on \dsetName{} is provided in Appendix Table \ref{tbl:model-info}. The exact prompt used to elicit LLM responses for evaluation under the basic prompt regime is provided in Appendix Figure \ref{fig:basic-prompt}. Under the expert-guided prompt regime, models were first instructed to generate a formatted article summmary using the summarization step (using Appendix Figure \ref{fig:expert-prompt:summarize}), then asked to provide answers based on the generated summaries for all relevant articles (via Appendix Figure \ref{fig:expert-prompt:answer}). In all cases, chunks of original article text or previously-generated summarization were provided with a header line containing the article's title, date of publication (if available), and PubMed ID,  allowing the LLM to recognize and assign blocks of content to different sources and synthesize in-context.

\begin{table}[h]
\centering
\caption{List of evaluated models with their model size and context length limit we set for our experiments. Precision is 16-bit floating point unless specified otherwise.}
\resizebox{\textwidth}{!}{
\begin{tabular}{llll}
\toprule
Model                                    & Model Type                    & Parameter Sizes          & Context Limit \\
\midrule
DeepSeek R1 \cite{deepseekR1}            & Generalist Reasoning          & 671B & 131K                    \\
DeepSeek V3 \cite{deepseekv3}            & Generalist Non-Reasoning      & 671B & 131K                    \\
GPT-4.1 \cite{gpt4}                      & Generalist Non-Reasoning      & Unknown & 1M                 \\
GPT-4.1 mini \cite{gpt4}                 & Generalist Non-Reasoning      & Unknown & 131K                 \\
GPT-o1 \cite{gpto1}                      & Generalist Non-Reasoning      & Unknown & 150K                 \\
HuatuoGPT-o1 \cite{huatuogpt-o1}         & Medical Reasoning             & 7B, 70B & 32K, 16K                 \\
Llama 3.0 \cite{llama3}                  & Generalist Non-Reasoning      & 8B, 70B & 8K                 \\
Llama 3.1 \cite{llama3}                  & Generalist Non-Reasoning      & 8B, 70B, 405B & 131K           \\
Llama 3.3 \cite{llama3}                  & Generalist Non-Reasoning      & 70B & 131K                      \\
Llama 3.3 (R1-Distill) \cite{deepseekR1} & Generalist Reasoning          & 70B & 131K                     \\
Llama 4 Maverick \cite{llama4}           & Generalist Non-Reasoning      & 400B (17B active) & 500K       \\
Llama 4 Scout \cite{llama4}              & Generalist Non-Reasoning      & 109B (17B active) & 1M       \\
OpenBioLLM \cite{OpenBioLLMs}            & Medical Non-Reasoning         & 8B, 70B  & 8K                \\
OpenThinker2 \cite{openthoughts}         & Generalist Reasoning          & 32B  & 131K                    \\
Qwen2.5 \cite{qwen2_5}                   & Generalist Non-Reasoning      & 7B, 32B, 72B  & 32K           \\
Qwen3 \cite{qwen3}                       & Generalist Reasoning (hybrid) & 235B (22B active, 8-bit) & 32 K\\
QwQ \cite{qwq32b}                        & Generalist Reasoning          & 32B   & 131K                   \\
\bottomrule
\end{tabular}
}
\label{tbl:model-info}
\end{table}

\begin{figure}[h]
    \begin{boxedminipage}[c]{\textwidth}
        \begin{lstlisting}
Given the ARTICLE SUMMARIES. Provide a concise and precise answer to the provided QUESTION.

After you think, return your answer with the following format:
- **Rationale**: Your rationale
- **Full Answer**:  A precise answer, citing each fact with the Article ID in brackets (e.g. [2]).
- **Answer**: A final classification exactly matching one of the following options: Higher, Lower, No Difference, Insufficient Data, Uncertain Effect

Think step by step.
**QUESTION**: {question}
**ARTICLE SUMMARIES**: {context}
        \end{lstlisting}
    \end{boxedminipage}
    \caption{Prompt used to generate LLM responses to questions under the basic prompt setup.}
    \label{fig:basic-prompt}
\end{figure}

\begin{figure}[h]
    \begin{subfigure}{\textwidth}
        \begin{boxedminipage}[c]{\textwidth}
            \begin{lstlisting}
You are the author of a Cochrane Collaboration systematic review, leveraging statistical analysis and assessing risks of bias in order to rigorously assess the effectiveness of medical interventions. As part of your review process, perform the following task:
As a subject expert, (1) summarize the evidence provided by a given ARTICLE as it pertains to a given QUESTION and (2) provide a possible answer.

Otherwise, if the provided article contains relevant information, you must return a list including the following items:

- **Study Design**: Type of study, level of evidence, and grade of recommendation according to the levels of evidence REC TABLE (provided Below).  
- **Study Population**: Study size and patient population.
- **Summary**: A concise but comprehensive summary based on the previously specified information, with a focus on the main findings.
- **Possible Answer**: A concise feasible answer given the evidence.

**REC TABLE **: Levels of Evidence (from strongest [1a] to lowest [5]).  

| Grade of Recommendation | Level of Evidence | Type of Study |  
|--------------|--------------|----------------------------------------------|
| A | 1a | Systematic review and meta-analysis of (homogeneous) randomized controlled trials |  
| A | 1b | Individual randomized controlled trials (with narrow confidence intervals) |  
| B | 2a | Systematic review of (homogeneous) cohort studies of `exposed' and `unexposed' subjects |  
| B | 2b | Individual cohort study / low-quality randomized control studies |  
| B | 3a | Systematic review of (homogeneous) case-control studies |  
| B | 3b | Individual case-control studies |  
| C | 4 | Case series, low-quality cohort or case-control studies, or case reports |  
| D | 5 | Expert opinions based on non-systematic reviews of results or mechanistic studies |" 

Think step by step.
**QUESTION**: {question}
**ARTICLE TITLE**: {title}
**ARTICLE CONTENT**:
{context}
            \end{lstlisting}
        \end{boxedminipage}
        \subcaption{Prompt used for the summarization step.}
        \label{fig:expert-prompt:summarize}
    \end{subfigure}
    \begin{subfigure}{\textwidth}
        \begin{boxedminipage}[c]{\textwidth}
            \begin{lstlisting}
You are the author of a Cochrane Collaboration systematic review, leveraging statistical analysis and assessing risks of bias in order to rigorously assess the effectiveness of medical interventions. As part of your review process, perform the following task:
Given the ARTICLE SUMMARIES. Provide a concise and precise answer to the provided QUESTION.

After you think, return your answer with the following format:
- **Rationale**: Your rationale
- **Full Answer**:  A precise answer, citing each fact with the Article ID in brackets (e.g. [2]).
- **Answer**: A final classification exactly matching one of the following options: Higher, Lower, No Difference, Insufficient Data, Uncertain Effect

Think step by step.
**QUESTION**: {question}
**ARTICLE SUMMARIES**: {context}
            \end{lstlisting}
        \end{boxedminipage}
        \subcaption{Prompt used for the final answer step.}
        \label{fig:expert-prompt:answer}
    \end{subfigure}
    \caption{Prompts used to generate LLM responses to questions under the expert-guided prompt setup, designed to attempt to explicitly enforce model awareness of evidence quality and strength.}
    \label{fig:expert-prompt}
\end{figure}

\FloatBarrier

\section{LLM instruction-following rates}
\label{a:sec:instr-follow}

The rate at which LLMs provided valid answer output of any kind is presented as part of Figure \ref{fig:base-acc_avg-perclass-recall}. Precisely, we measured the per-model instruction-following rate, i.e. the percentage of questions for which the full ``Answer" field in the model's final output exactly matched one of the defined answer classes (case-insensitive). We note that a substantial portion of models exhibit a high rate of instruction-following failures: OpenBioLLM 8B and 70B; HuatuoGPT-o1 7B and 70B; Llama 4 Maverick and Scout; Llama 3.0 8B; and Llama 3.1 8B all fail to achieve a $60$\% instruction-following rate, and only Llama 3.3 70B (Instruct and R1-Distill) achieves perfect instruction-following. We highlight that OpenBioLLM 8B has a $0$\% instruction-following rate. Lastly, we observe that even when  significant portion of the  outputs are valid, models  still have high error rates, with only an average of $58.1 (\pm 5.0)$\% of valid model outputs being correct. These results demonstrate that, while a high instruction-following rate may diminish performance in small models, poor performance cannot be attributed to instruction-following errors alone.

\section{LLM performance as a function of number of relevant sources}
As shown in Appendix Figure \ref{fig:num-sources}, we find no clear general trend between the number of relevant sources and model performance. Notably, this includes performance with a single source (no model achieves even $60$\% accuracy), highlighting challenges in LLMs' ability to perform systematic review beyond resolving evidence conflicts. The only exceptions to this are the models with the overall poorest performance (colored in red and orange hues, such as HuatuoGPT-o1 7B and Llama 3.0 8B).
\begin{figure}[h!]
    \centering
    \includegraphics[width=0.9\linewidth]{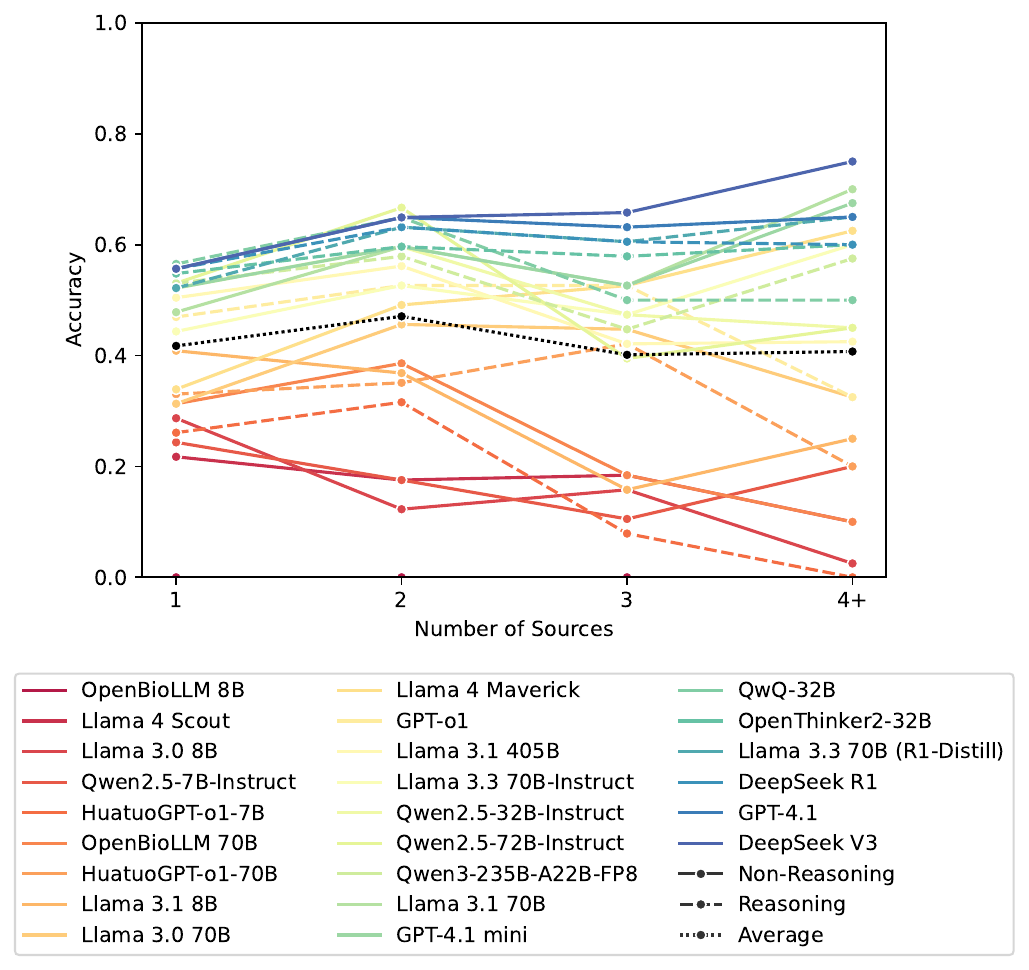}
    \caption{Model accuracy as a function of number of relevant sources.}
    \label{fig:num-sources}
\end{figure}

\FloatBarrier

\section{LLM performance as a function of token length of relevant sources}

Given the lack of dependency on the number of sources on average accuracy, we directly investigate the dependency of model performance on the combined token length of all relevant sources; we present these results in Appendix Figure \ref{fig:num-tokens}. As noted in the main analysis, performance consistently declines at high token counts, except for models with over 100B parameters. Notably, 32B models maintain over 50\% average accuracy up to the 80–100\% quantile (15K tokens and above). By contrast, 70–72B models fall below 50\% accuracy around the 60–80\% quantile (11–15K tokens). This decline in the 70–72B range is primarily driven by the underperformance of medically finetuned models (HuatuoGPT-o1 and OpenBioLLM). 
\begin{figure}
    \centering
    \includegraphics[width=1.0\linewidth]{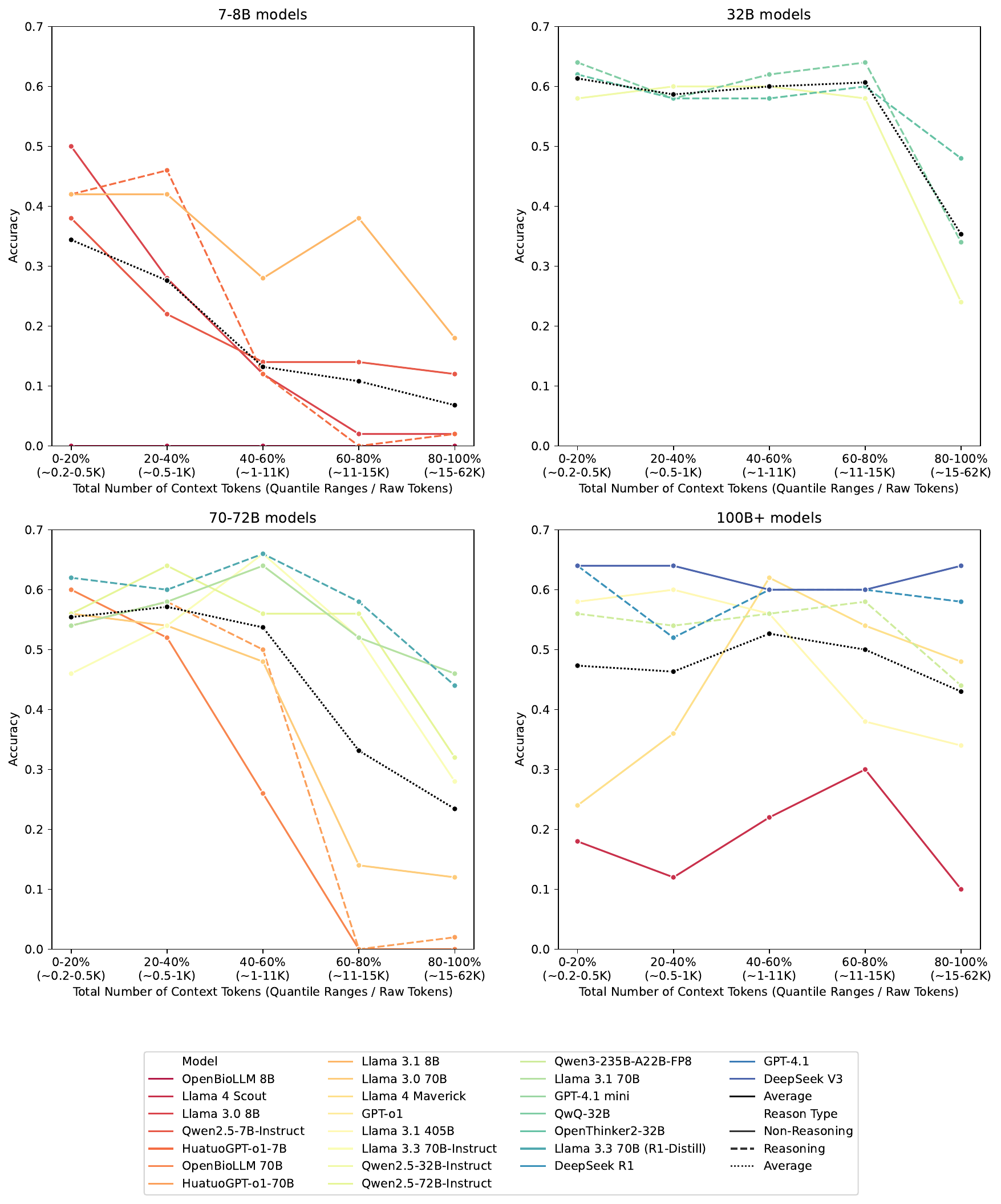}
    \caption{Model performance as a function of the number of tokens in the relevant studies, separated by model size range. Horizontal axis measures the accuracy by 5-quantiles.}
    \label{fig:num-tokens}
\end{figure}

\FloatBarrier

\section{Average confusion matrices for treatment outcome effects}

\begin{figure}
    \centering
    \makebox[\textwidth][c]{\includegraphics[width=1.3\textwidth]{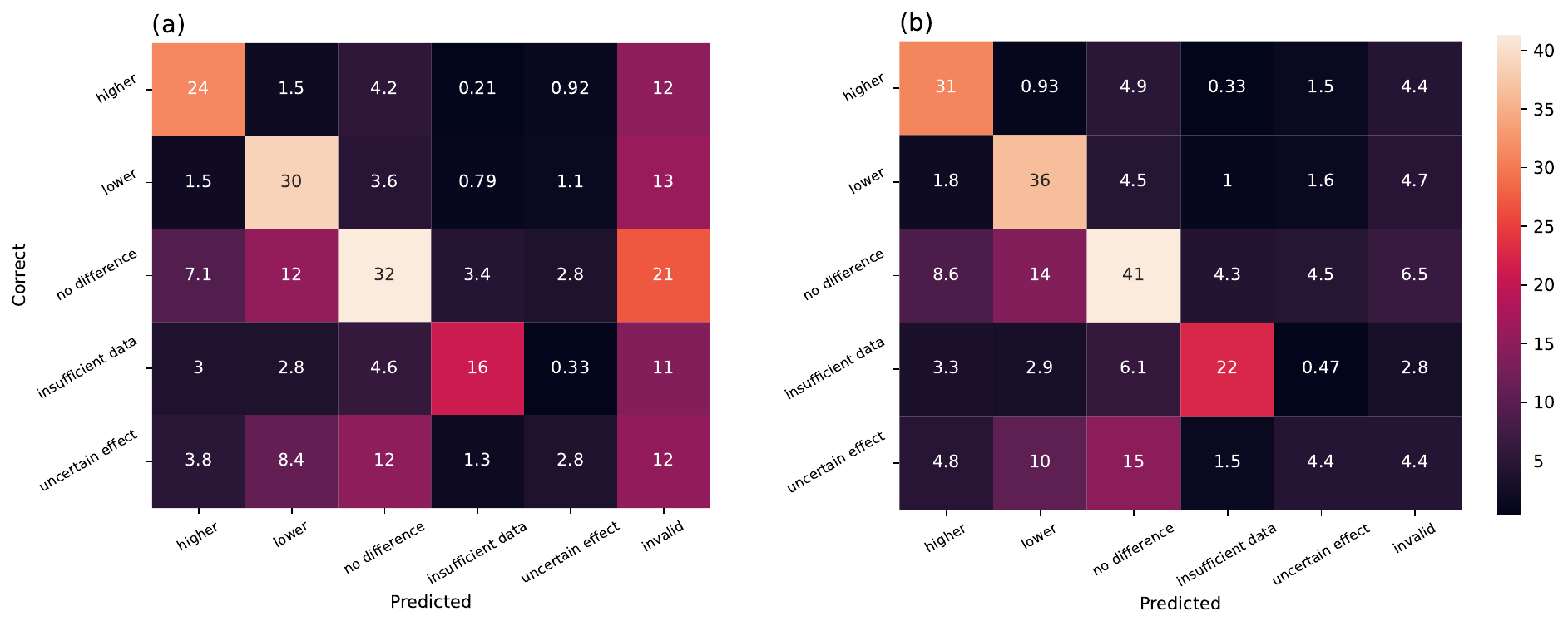}}
    \caption{Average confusion matrices using basic prompts. (a) Average confusion matrix aggregated across all models. (b) Average confusion matrix aggregated across models achieving at least 40\% overall accuracy.}
    \label{fig:avg-confusion-matrix}
\end{figure}
We assess which treatment outcome effect classes are most frequently misclassify  by visualizing the confusion matrix averaged across all models. As shown in  Figure~\ref{fig:avg-confusion-matrix}, we observe that models with lower than 40\% accuracy significantly skew the confusion matrix toward invalid outputs. However, when considering  exclusively models with above 40\% performance, we observe two significant trends. First, models are consistently unwilling to predict \texttt{uncertain effect}. 
 Second, models consistently confuse the \texttt{uncertain effect} and \texttt{no difference} classes. 

For completeness, we provide all individual confusion matrices in Appendix Section \ref{a:sec:allConfMats}.

\section{Performance by review publication year}
\begin{figure}
    \centering
    \includegraphics[width=\linewidth]{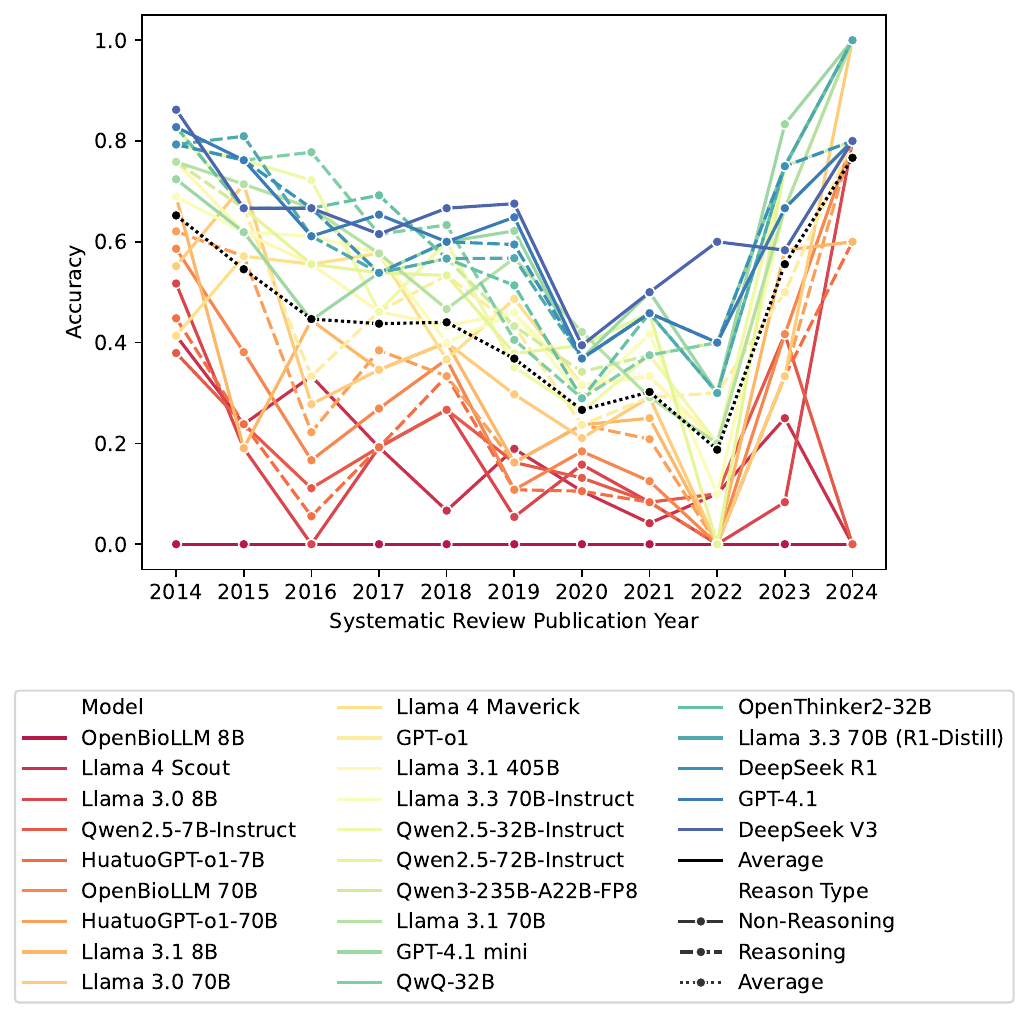}
    \caption{Accuracy by publication year}
    \label{fig:pubyear-acc}
\end{figure}
As shown in Appendix Figure~\ref{fig:pubyear-acc}, performance steadily declines for more recent publication years, except for 2023 and 2024. These improvements may partially be explained by the fact that the majority of questions from 2024 involve high- or moderate-certainty evidence (as shown in Appendix Figure~\ref{fig:extra-data-distrib}(d)); as a result, these questions are likely easier for models to answer.

\section{Per-class recall for individual models}
\begin{figure}
    \makebox[\textwidth][c]{\includegraphics[width=1.5\textwidth]{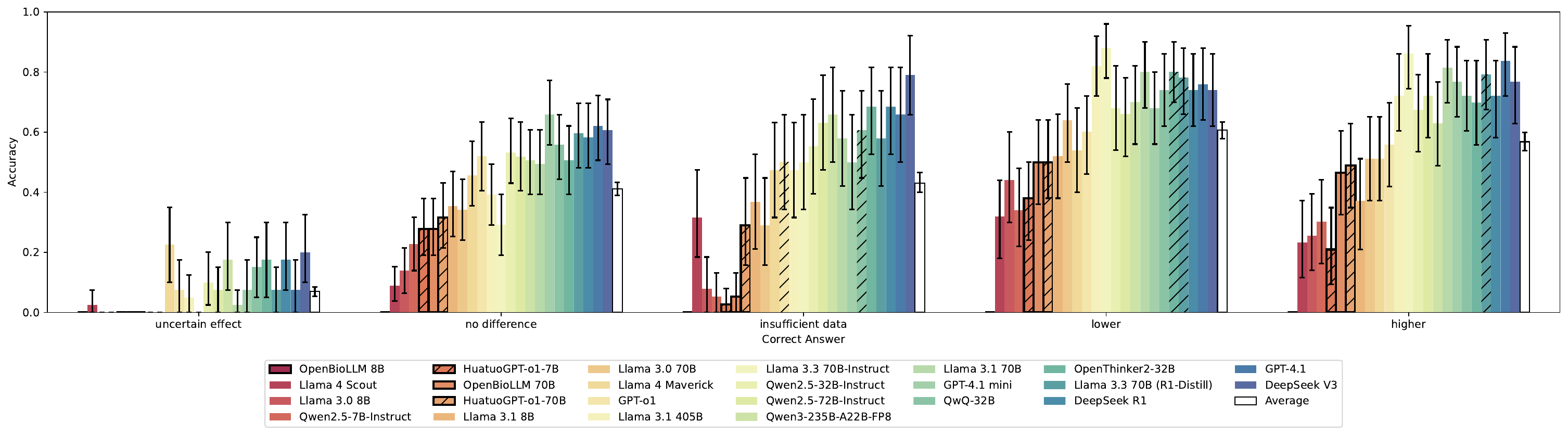}}
    \caption{Per-class recall for each individual model. 95\% confidence intervals are calculated via bootstrapping with N=1000.}
    \label{fig:per-class-recall}
\end{figure}
We present individual model per-class recall in Appendix Figure \ref{fig:per-class-recall}. Notably, all models, without exception, perform poorly on the \texttt{uncertain effect} class. We highlight that Llama 3.3 70B-Instruct outperforms all other models on the \texttt{higher} and \texttt{lower} classes, but its overall accuracy is held back significantly by its poor performance on the \texttt{no difference} and \texttt{insufficient data} classes.

\FloatBarrier

\section{Model performance under the expert-guided prompt setup}
\begin{figure}[b!]
    \centering
    \includegraphics[width=\linewidth]{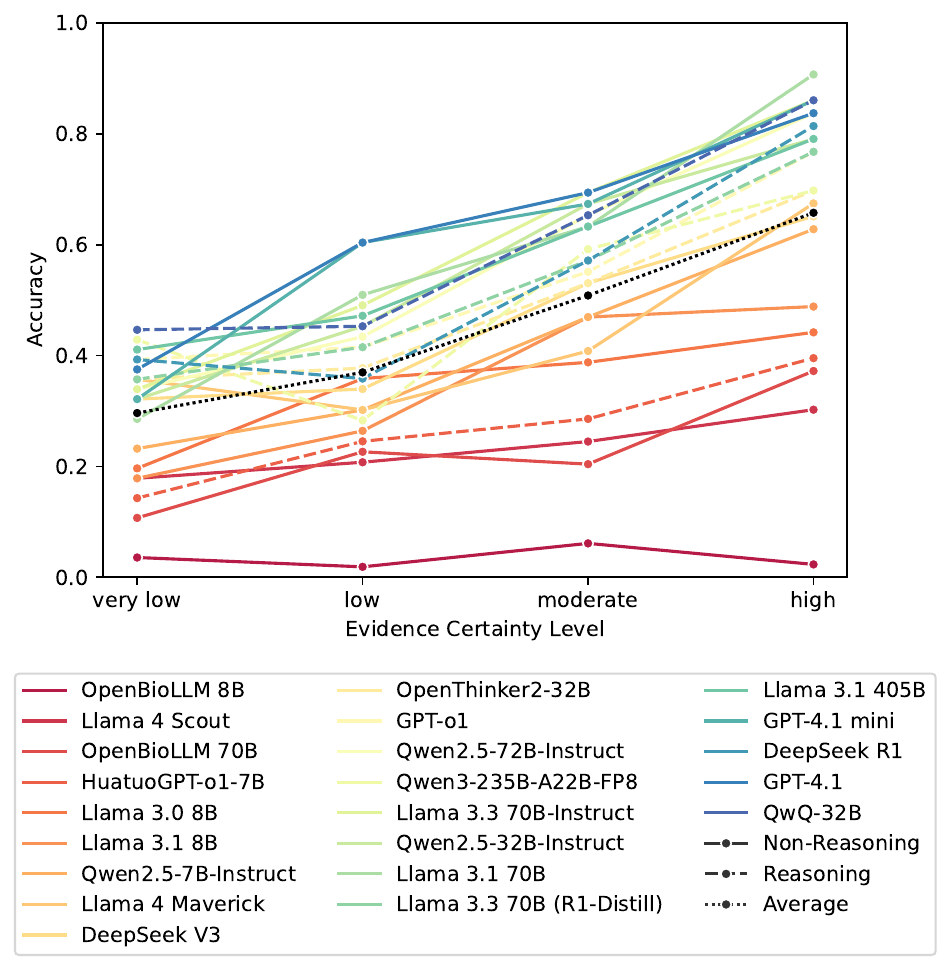}
    \caption{Model accuracy at different evidence qualities when using the expert-guided prompt setup. HuatuoGPT-o1 70B and Llama 3.0 70B are omitted as they were not tested on the expert-guided setup.}
    \label{fig:expert-certainty}
\end{figure}
\begin{figure}
    \centering
    \includegraphics[width=0.8\linewidth]{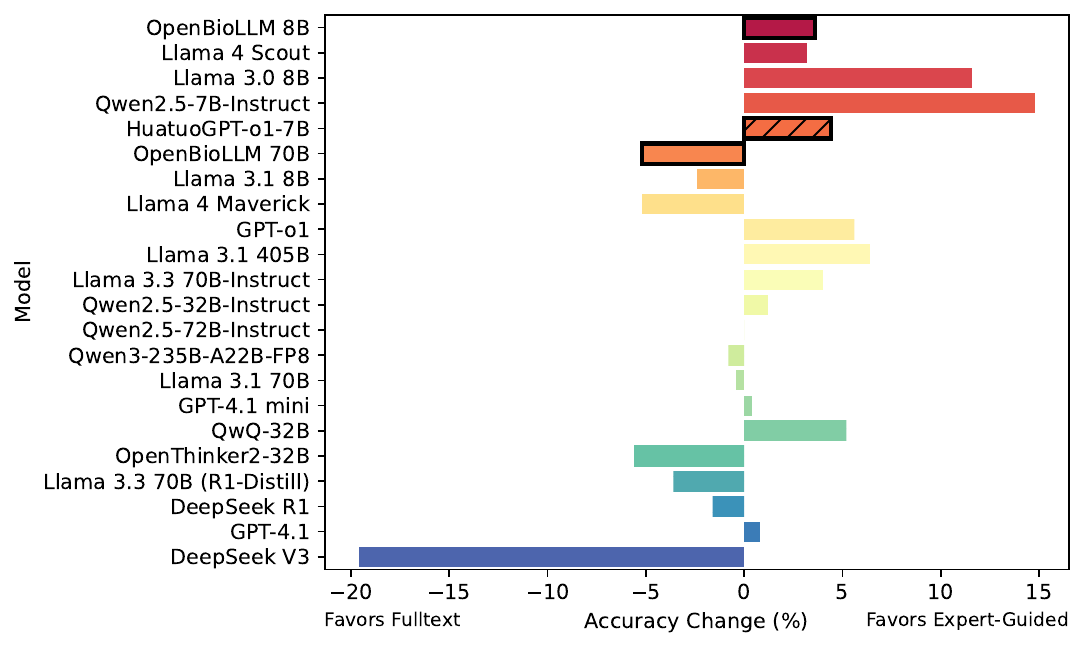}
    \caption{Changes in model performance when using the basic prompt setup versus the expert-guided prompt setup. HuatuoGPT-o1 70B and Llama 3.0 70B are omitted as they were not tested on the expert-guided setup.}
    \label{fig:basic_vs_expguided}
\end{figure}
To evaluate the dependency of model performance on prompting quality, we leverage an expert-guided prompt setup as described in the main paper and Appendix Section~\ref{a:sec:models_prompts}. Critically, as shown in Appendix Figure \ref{fig:expert-certainty} and discussed in the main paper, we find that even with a prompt explicitly designed to encourage models to assess the quality of studies, the dependency of model performance on evidence certainty remains. More broadly, as shown in Appendix Figure~\ref{fig:basic_vs_expguided}, we find that our more intentionally-designed prompt does not consistently improve model performance; while performance improves for the five models that performed worst under the basic prompt (namely OpenBioLLM 8B, Llama 4 Scout, Llama 3.0 8B, Qwen2.5-7B-Instruct, and HuatuoGPT-o1 7B), we observe that performance actually decreases for several of the models that performed best with the basic prompt, including a nearly 20\% drop in performance for DeepSeek V3 (the highest-performing model when using the basic prompt).

\FloatBarrier

\section{Question correctness across models}
\begin{figure}
    \centering
    \includegraphics[width=\linewidth]{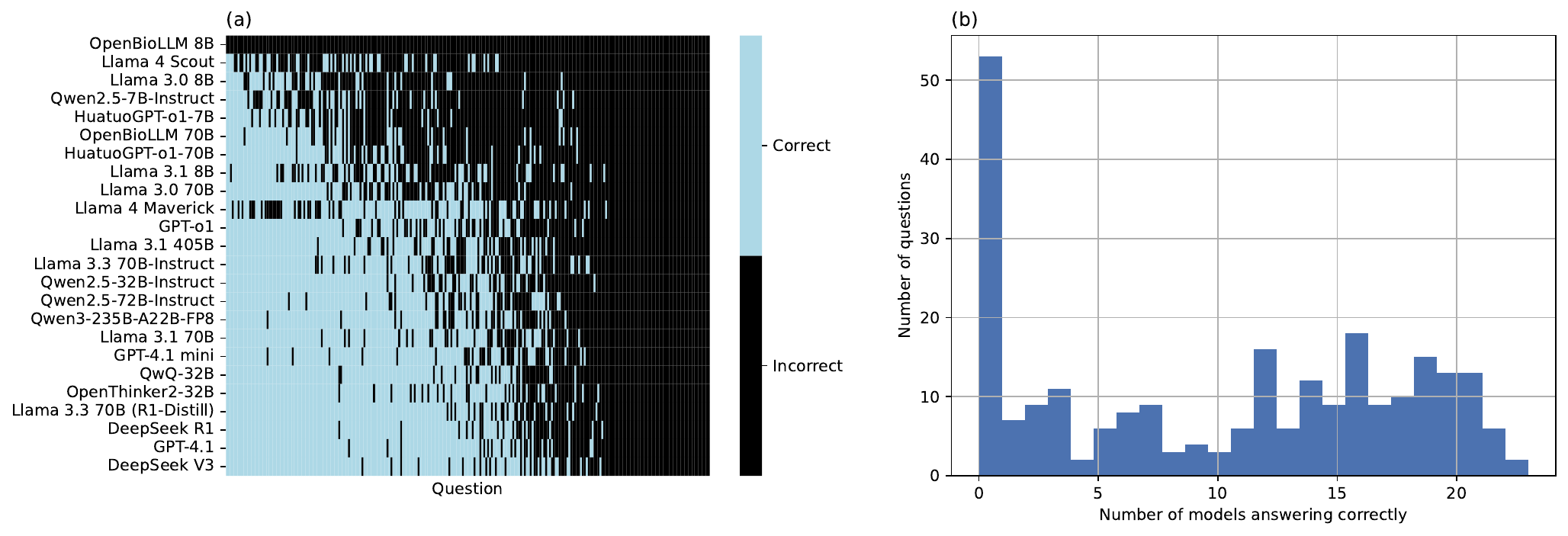}
    \caption{Analyses of model behavior across questions. (a) Questions (columns) that were deemed correct (light blue) or incorrect (black) for each model (rows), sorted by percentage of models with correct responses for that question (x-axis) and by the percentage of questions a model got correct (y-axis). (b) Distribution of questions by the number of models that answered that question correctly.}
    \label{fig:q-acc-by-model}
\end{figure}
As shown in Appendix Figure \ref{fig:q-acc-by-model}, 53 questions are answered incorrectly by all models, and only 2 are answered correctly by all models (omitting OpenBioLLM 8B, which gets every question wrong). Otherwise, we observe that performance varies significantly across models . A qualitative analysis of these various question types is presented in Appendix Section~\ref{a:sec:qualitative_analysis}.

\section{Performance by medical specialty}
\begin{figure}
    \centering
    \includegraphics[width=0.8\linewidth]{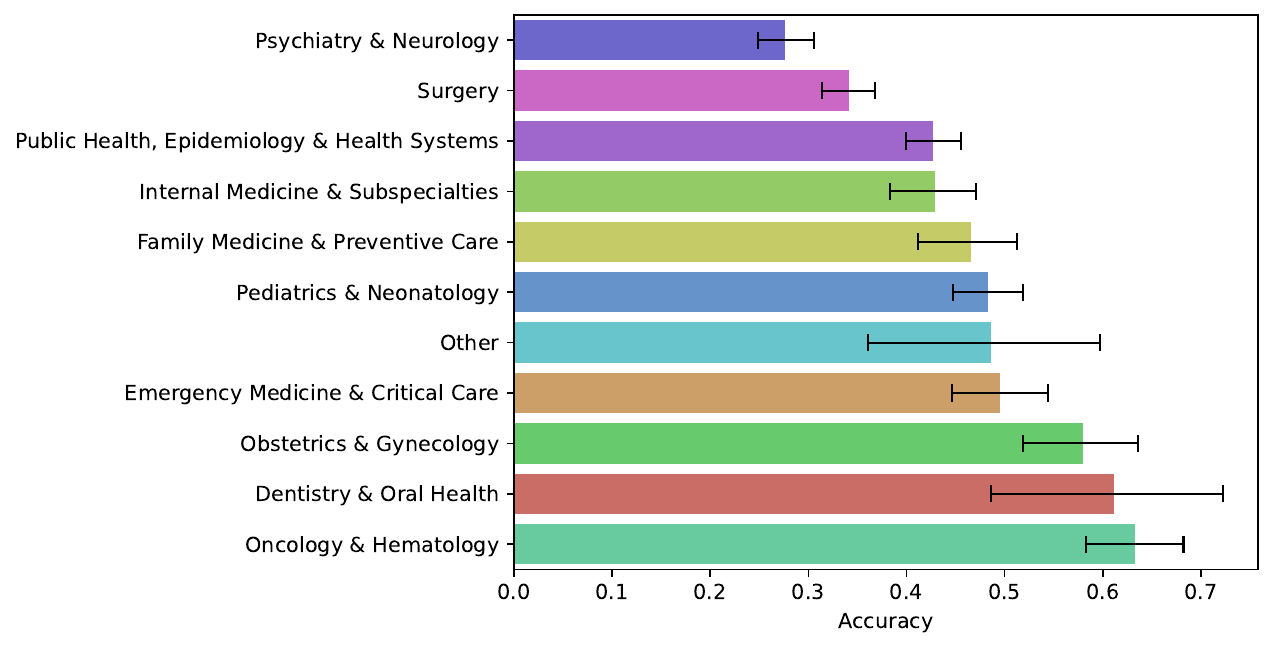}
    \caption{Average model accuracy across all models (and 95\% confidence interval) stratified by medical specialty.}
    \label{fig:med_specialty_acc}
\end{figure}
 Appendix Figure \ref{fig:med_specialty_acc} shows average model accuracy stratified by medical specialty. Models perform significantly worse on questions relating to Psychology \& Neurology and Surgery relative to other medical specialties, with accuracies of 27.60\% (24.58, 30.52) and 34.09\% (31.15, 37.03) respectively. The highest average model performance is observed in the Oncology \& Hematology specialty, where models achieve an average accuracy of 63.28\% (95\% CI: 58.33–68.23).

\section{Full-text vs abstract sources}
\begin{figure}
    \centering
    \includegraphics[width=0.8\linewidth]{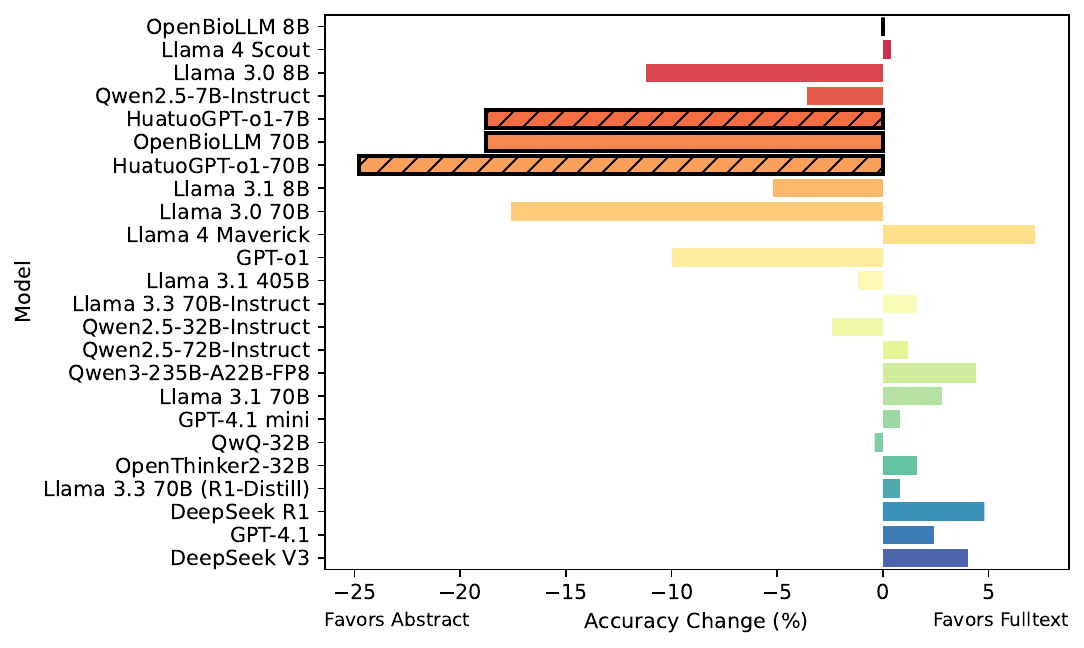}
    \caption{Changes in model performance when providing full-text when available versus always providing only the abstract (basic prompt setup).}
    \label{fig:fulltext_vs_abstract}
\end{figure}
We evaluate how model performance differs when using full-text articles versus abstracts alone, using the basic prompt setup in both cases.
The results of this experiment are presented in Appendix Figure \ref{fig:fulltext_vs_abstract}. We find that most models with the poorest overall performance actually experience a boost in accuracy (as high as 24.8\% in the case of HuatuoGPT-o1-70B) when given only abstracts, even though abstracts contain less information. This suggests that some models struggle on our dataset because of an inability to handle long context, as full-text articles are much longer than abstracts alone. By contrast, the best-performing models usually perform better when given access to full-text (DeepSeek R1, for instance, gains 4.8\% accuracy with full-text access). We note that, due to licensing constraints in scientific publishing, many existing deployments and evaluations of LLM to answer questions using scientific literature rely solely on abstracts  \cite{lozano2023clinfo,openevidence2025}. Our analysis demonstrates that access to full article text benefits frontier models, underscoring the urgent need to expand such access. We highlight initiatives promoting this shift—for instance, beginning in 2025, all U.S. federally-funded research must be made freely available, which could significantly enhance the performance of already-deployed models.

\FloatBarrier
 
\section{Qualitative analysis of DeepSeek V3}
\label{a:sec:qualitative_analysis}
To provide a better qualitative understanding of success and failure modes on our dataset, we explore the outputs (under the basic prompt setup) of our best-performing model, DeepSeek V3, and provide some commentary on the quality of the model's output. Specifically, we consider four types of questions: (1) Questions where all models get the answer wrong. (2) Questions where most models are correct, but DeepSeek V3 is incorrect. (3) Questions where most models are correct, including DeepSeek V3. (4) Questions where most models are wrong, but DeepSeek V3 is correct. 

\subsection{Questions where all models are wrong}

\begin{QualExample}{purple}{Question ID 1 - 0/24 models correct}
    \textbf{Review PMID:} 28660687\\
    \textbf{Question:} Is the short-term rate of overall lymphocyst formation higher, lower, or the same when comparing retroperitoneal drainage to no drainage?\\
    \textbf{Correct Answer:} \texttt{no difference}\\
    \textbf{PMIDs of Relevant Sources:} 9190979, 12214830\\
    \noindent\rule{\linewidth}{0.4pt}\\
    \textbf{DeepSeek V3 Response:}
    \begin{lstlisting}
- **Rationale**: 
  - Article ID 9190979 reports that the complication rate, including lymphocyst formation, was significantly higher in the drained group (Group A) compared to the not-drained group (Group B) (P = 0.01). However, sonographic monitoring showed free abdominal fluid (a proxy for lymphocyst formation) in 18% of drained patients vs. 36% of not-drained patients (P = 0.03), suggesting a lower rate in the drained group. This creates a contradiction in the findings.
  - Article ID 12214830 found no significant difference in lymphocyst formation between the drained and not-drained groups (P = 0.2), with asymptomatic lymphocysts detected only in the not-drained group (though statistically insignificant).

- **Full Answer**: 
  - Article ID 9190979 reports a higher complication rate (including lymphocyst-related morbidity) in the drained group [9190979], but sonographic data suggests a lower rate of free abdominal fluid (a proxy for lymphocyst formation) in the drained group (18% vs. 36%, P = 0.03) [9190979]. Article ID 12214830 found no significant difference in lymphocyst formation between groups (P = 0.2) [12214830]. The evidence is conflicting, with one study suggesting a lower rate and another showing no difference.

- **Answer**: Uncertain Effect
    \end{lstlisting}
    \noindent\rule{\linewidth}{0.4pt}
    \textbf{Remarks:} In the SR, authors perform meta-analysis and find no overall significant difference between retroperitoneal drainage and no drainage. DeepSeek V3 does not appear to know how to perform any kind of reconcilliation. We note that Article 12214830 does not specify the raw number of events under both treatments, but does state that no lymphocysts (i.e. 0) were detected in the drainage group.
\end{QualExample}

\begin{QualExample}{purple}{Question ID 198 - 0/24 models correct}
    \textbf{Review PMID:} 35514111\\
    \textbf{Question:} Is the rate of any PCR‐positive SARS‐COV‐2 infection higher, lower, or the same when comparing test‐based attendance to standard 10‐day self‐isolation?\\
    \textbf{Correct Answer:} \texttt{uncertain effect}\\
    \textbf{PMIDs of Relevant Sources:} 34534517\\
    \noindent\rule{\linewidth}{0.4pt}\\
    \textbf{DeepSeek V3 Response:}
    \begin{lstlisting}
- **Rationale**: The study compared daily testing (test-based attendance) with standard 10-day self-isolation for school-based contacts of SARS-CoV-2 cases. The primary outcomes included rates of symptomatic PCR-confirmed infections and COVID-19-related absences. The results showed no significant difference in the rate of symptomatic PCR-confirmed infections between the two groups, with the intervention group (daily testing) being non-inferior to the control group (self-isolation). The adjusted incidence rate ratio (aIRR) for symptomatic PCR-confirmed infections was 0.96 (95% CI 0.75–1.22), indicating no statistically significant difference.  

- **Full Answer**: The rate of PCR-positive SARS-CoV-2 infection was not significantly different when comparing test-based attendance (daily testing) to standard 10-day self-isolation (aIRR 0.96, 95% CI 0.75–1.22) [34534517].  

- **Answer**: No Difference
    \end{lstlisting}
    \noindent\rule{\linewidth}{0.4pt}
    \textbf{Remarks:} The authors of the SR state that the results are inconclusive for the following reasons: (a) "risk of bias: considerable missing data for PCR-tests for staff (data were available for 76\% control schools and 83\% intervention schools)" and (b) "imprecision: the confidence intervals are wide and the 95\% CI crosses the line of no effect." In this case, the full-text article is open-source, so LLMs theoretically had access to essentially all the same information as the SR authors, but evidently, none of these concerns are addressed in DeepSeek V3's response. Furthermore, the DeepSeek V3 response focuses only on symptomatic PCR-confirmed infections, demonstrating limitations in its knowledge of medical contexts, as the question calls for an assessment of any PCR positive infections, which includes asymptomatic cases.
\end{QualExample}

\begin{QualExample}{purple}{Question ID 32 - 0/24 models correct}
    \textbf{Review PMID:} 26447539\\
    \textbf{Question:} Is subjective fatigue higher, lower, or the same when comparing methylphenidate to placebo?\\
    \textbf{Correct Answer:} \texttt{no difference}\\
    \textbf{PMIDs of Relevant Sources:} 17674415\\
    \noindent\rule{\linewidth}{0.4pt}\\
    \textbf{DeepSeek V3 Response:}
    \begin{lstlisting}
- **Rationale**: The study (Article ID 17674415) directly compares methylphenidate to placebo in treating fatigue in Parkinson's disease patients. The results show that methylphenidate significantly reduced fatigue scores (FSS and MFI) compared to placebo, where reductions were nonsignificant. This indicates that subjective fatigue is lower with methylphenidate than with placebo.

- **Full Answer**: Subjective fatigue is lower with methylphenidate compared to placebo, as evidenced by significant reductions in Fatigue Severity Scale (FSS) and Multidimensional Fatigue Inventory (MFI) scores in the methylphenidate group (P < 0.04), while placebo group reductions were nonsignificant [17674415].

- **Answer**: Lower
    \end{lstlisting}
    \noindent\rule{\linewidth}{0.4pt}
    \textbf{Remarks:} This is a relatively challenging question—in the original SR, the authors compare the improvements in FSS in the placebo group directly to the improvements in the treatment group, rather than comparing the two treatment groups separately to their own baselines (which is what the authors of the cited source do). The source's abstract does state that smaller but non-significant improvements were present in the placebo group, albeit without providing numbers. Regardless, the LLM's response reads as a regurgitation of facts and lacks the critical thinking needed to critically assess cited studies.
\end{QualExample}

\subsection{Questions where most models are correct, but DeepSeek V3 is wrong}

\begin{QualExample}{purple}{Question ID 6 - 16/24 models correct}
    \textbf{Review PMID:} 26346232\\
    \textbf{Question:} Is stroke prevention higher, lower, or the same when comparing Transcatheter Device Closure (TDC) to medical therapy?\\
    \textbf{Correct Answer:} \texttt{no difference}\\
    \textbf{PMIDs of Relevant Sources:} 22417252, 23514285, 23514286\\
    \noindent\rule{\linewidth}{0.4pt}\\
    \textbf{DeepSeek V3 Response:}
    \begin{lstlisting}
- **Rationale**: The three articles provide mixed evidence on the effectiveness of Transcatheter Device Closure (TDC) compared to medical therapy for stroke prevention in patients with patent foramen ovale (PFO). Article ID 22417252 found no significant difference in stroke prevention between TDC and medical therapy (P=0.37 for the primary endpoint). Article ID 23514285 also reported no significant reduction in recurrent embolic events or death with TDC (P=0.34 for the primary endpoint). However, Article ID 23514286 showed no significant benefit in the intention-to-treat analysis (P=0.08) but did find a significant reduction in recurrent stroke in the per-protocol and as-treated analyses (P=0.03 and P=0.007, respectively). Given the inconsistency in results across studies and analyses, the overall evidence is not conclusive.

- **Full Answer**: The evidence is mixed. Two studies (Article ID 22417252 and Article ID 23514285) found no significant difference in stroke prevention between TDC and medical therapy [22417252][23514285]. A third study (Article ID 23514286) reported no significant benefit in the intention-to-treat analysis but found TDC superior in per-protocol and as-treated analyses [23514286]. 

- **Answer**: Uncertain Effect
    \end{lstlisting}
    \noindent\rule{\linewidth}{0.4pt}
    \textbf{Remarks:} In this case, unlike most models (16/24 to be exact), DeepSeek V3 seems to 'overthink' the question and hone in on a single case where a sub-analysis of in Article 23514286 finds a significant benefit. Notably, SRs typically use intention-to-treat analysis (https://training.cochrane.org/handbook/current/chapter-06), which DeepSeek V3 fails to consider. Furthermore, as in prevous cases, no clear efforts to perform meta-analysis or aggregate studies are shown in DeepSeek V3's response.
\end{QualExample}

\begin{QualExample}{purple}{Question ID 106 - 15/24 models correct}
    \textbf{Review PMID:} 38197546\\
    \textbf{Question:} Is the risk of cholera at two-year follow-up higher, lower, or the same when comparing two doses of Dukoral with or without a booster dose to placebo?\\
    \textbf{Correct Answer:} \texttt{lower}\\
    \textbf{PMIDs of Relevant Sources:} 7967990, 10823767\\
    \noindent\rule{\linewidth}{0.4pt}\\
    \textbf{DeepSeek V3 Response:}
    \begin{lstlisting}
- **Rationale**: 
  - Article ID 7967990 evaluates the short-term efficacy of two doses of WC/rBS cholera vaccine without a booster, showing 86% protective efficacy over a mean follow-up of 18 weeks. However, it does not provide data on long-term (two-year) efficacy or the effect of a booster dose.
  - Article ID 10823767 evaluates the two-year efficacy of the same vaccine, comparing two doses (without booster) to three doses (with booster). After two doses, the vaccine showed no efficacy (VE = -4%), but after the booster dose, efficacy increased to 61%. This suggests that the booster dose significantly improves long-term protection compared to two doses alone.

- **Full Answer**: 
  - Without a booster dose (two doses only), the vaccine showed no protective efficacy (VE = -4%) at two-year follow-up [10823767]. 
  - With a booster dose (three doses), the vaccine demonstrated 61% protective efficacy at two-year follow-up [10823767]. 
  - Therefore, the risk of cholera at two-year follow-up is **higher** with two doses alone compared to two doses plus a booster.

- **Answer**: Higher
    \end{lstlisting}
    \noindent\rule{\linewidth}{0.4pt}
    \textbf{Remarks:} This is case where DeepSeek V3 seems to fail due to a misinterpretation of the question, although we believe the semantics of this question are unambiguous. DeepSeek V3 does not seem to pick up on the mention of placebo in the question at all.
\end{QualExample}

\begin{QualExample}{purple}{Question ID 163 - 16/24 models correct}
    \textbf{Review PMID:} 25739381\\
    \textbf{Question:} Is neonatal mortality higher, lower, or the same when comparing community-led chlorhexidine cord cleansing to community-led dry cord care?\\
    \textbf{Correct Answer:} \texttt{lower}\\
    \textbf{PMIDs of Relevant Sources:} 22322124, 16546539, 22322126\\
    \noindent\rule{\linewidth}{0.4pt}\\
    \textbf{DeepSeek V3 Response:}
    \begin{lstlisting}
- **Rationale**: The question asks whether neonatal mortality is higher, lower, or the same when comparing community-led chlorhexidine cord cleansing to community-led dry cord care. The three articles provide evidence from randomized trials in different settings (Bangladesh, Nepal, and Pakistan) comparing chlorhexidine cord cleansing with dry cord care. All three studies report reductions in neonatal mortality with chlorhexidine use, though the magnitude and statistical significance vary.  

- **Full Answer**:  
  - In Bangladesh, neonatal mortality was lower with single chlorhexidine cleansing (22.5 per 1000 livebirths) compared to dry cord care (28.3 per 1000 livebirths; RR 0.80 [0.65-0.98]) [22322124].  
  - In Nepal, neonatal mortality was 24% lower with chlorhexidine (RR 0.76 [0.55-1.04]) compared to dry cord care, with a significant 34% reduction in infants enrolled within 24 hours (RR 0.66 [0.46-0.95]) [16546539].  
  - In Pakistan, chlorhexidine cleansing reduced neonatal mortality (RR 0.62 [0.45-0.85]) compared to dry cord care [22322126].  

- **Answer**: **Lower**
    \end{lstlisting}
    \noindent\rule{\linewidth}{0.4pt}
    \textbf{Remarks:} An example of an instruction-following failure for DeepSeekV3 (even though the overall analysis leads to a correct solution).
\end{QualExample}

\subsection{Questions where most models are correct, including DeepSeek V3}

\begin{QualExample}{teal}{Question ID 160 - 21/24 models correct}
    \textbf{Review PMID:} 28829911\\
    \textbf{Question:} Is the risk of treatment‐related mortality higher, lower, or the same when comparing chemoradiotherapy plus esophagectomy to chemoradiotherapy alone?\\
    \textbf{Correct Answer:} \texttt{higher}\\
    \textbf{PMIDs of Relevant Sources:} 17401004, 15800321\\
    \noindent\rule{\linewidth}{0.4pt}\\
    \textbf{DeepSeek V3 Response:}
    \begin{lstlisting}
- **Rationale**: Both articles compare chemoradiotherapy plus esophagectomy (surgery) to chemoradiotherapy alone in patients with locally advanced esophageal cancer. Article ID 17401004 reports a 3-month mortality rate of 9.3% in the surgery group (arm A) compared to 0.8% in the chemoradiation-alone group (arm B), with a statistically significant difference (P = .002). Article ID 15800321 similarly reports a significantly higher treatment-related mortality in the surgery group (12.8%) compared to the chemoradiotherapy-alone group (3.5%) (P = .03). Both studies consistently demonstrate that adding surgery to chemoradiotherapy increases treatment-related mortality.

- **Full Answer**: The risk of treatment-related mortality is higher when comparing chemoradiotherapy plus esophagectomy to chemoradiotherapy alone, as evidenced by significantly higher mortality rates in the surgery groups (9.3% vs. 0.8% [17401004]; 12.8% vs. 3.5% [15800321]).

- **Answer**: Higher
    \end{lstlisting}
    \noindent\rule{\linewidth}{0.4pt}
    \textbf{Remarks:} While this question involves multiple sources, both sources individually agree with the overall correct conclusion, making it fairly straightforward to answer; consequently, almost all models get this question correct.
\end{QualExample}

\begin{QualExample}{teal}{Question ID 214 - 23/24 models correct}
    \textbf{Review PMID:} 37594020\\
    \textbf{Question:} Is pain relief at 48 hours higher, lower, or the same when comparing NSAIDs to placebo?\\
    \textbf{Correct Answer:} \texttt{higher}\\
    \textbf{PMIDs of Relevant Sources:} 8871138\\
    \noindent\rule{\linewidth}{0.4pt}\\
    \textbf{DeepSeek V3 Response:}
    \begin{lstlisting}
- **Rationale**: The study compared ibuprofen (an NSAID) to acetaminophen and placebo for pain relief in children with acute otitis media over 48 hours. The results showed that ibuprofen was statistically superior to placebo (P < 0.01) in relieving pain, with 7% of children still suffering pain in the ibuprofen group compared to 25% in the placebo group. There was no significant difference between acetaminophen and placebo for pain relief.  

- **Full Answer**: Pain relief at 48 hours was higher with NSAIDs (ibuprofen) compared to placebo, as evidenced by a statistically significant reduction in pain (7% vs. 25% of children still suffering pain, P < 0.01) [8871138].  

- **Answer**: Higher
    \end{lstlisting}
    \noindent\rule{\linewidth}{0.4pt}
    \textbf{Remarks:} A straightforward question that all models (except OpenBioLLM 8B) get correct: one source accurately directly reports on relevant findings without conclusions being revised by the SR authors' critical analysis.
\end{QualExample}

\subsection{Questions where DeepSeek V3 is correct, despite most models being wrong}

\begin{QualExample}{teal}{Question ID 235 - 3/24 models correct}
    \textbf{Review PMID:} 32104914\\
    \textbf{Question:} Is quality of life higher, lower, or the same when comparing computerised cognitive training (CCT) of at least 12 weeks to no intervention?\\
    \textbf{Correct Answer:} \texttt{insufficient data}\\
    \textbf{PMIDs of Relevant Sources:} 23531885, 25511081, 20418350\\
    \noindent\rule{\linewidth}{0.4pt}\\
    \textbf{DeepSeek V3 Response:}
    \begin{lstlisting}
- **Rationale**: The question asks whether quality of life (QoL) is higher, lower, or the same when comparing computerized cognitive training (CCT) of at least 12 weeks to no intervention. The provided articles focus on cognitive and physical performance improvements but do not explicitly measure or report QoL outcomes. Article ID 23531885 discusses cognitive improvements from CCT but does not mention QoL. Article ID 25511081 examines dual-task gait and divided attention but also does not address QoL. Article ID 20418350 evaluates cognitive performance after mental and physical activity interventions but similarly lacks QoL data.  

- **Full Answer**: None of the articles provide data on quality of life (QoL) outcomes when comparing CCT of at least 12 weeks to no intervention [23531885, 25511081, 20418350].  

- **Answer**: Insufficient Data
    \end{lstlisting}
    \noindent\rule{\linewidth}{0.4pt}
    \textbf{Remarks:} DeepSeek V3 correctly recognizes that none of the relevant studies provide direct, quantitative analysis on QoL outcomes; given the very low model correctness rate (3/24), other models likely attempted to apply colloquial definitions of "quality of life" rather than the narrowly-defined medical definition requiring explicit quantitative measurement.
\end{QualExample}

\begin{QualExample}{teal}{Question ID 89 - 2/24 models correct}
    \textbf{Review PMID:} 31557310\\
    \textbf{Question:} Is the incidence of surgical site infection (SSI) for patients undergoing breast cancer surgery without reconstruction higher, lower, or the same when comparing prophylactic antibiotics administered preoperatively to placebo?\\
    \textbf{Correct Answer:} \texttt{lower}\\
    \textbf{PMIDs of Relevant Sources:} 7480237, 9776150, 23052809, 10873356, 19673597, 2403655, 2403697\\
    \noindent\rule{\linewidth}{0.4pt}\\
    \textbf{DeepSeek V3 Response:}
    \begin{lstlisting}
- **Rationale**: The question asks about the incidence of surgical site infection (SSI) in breast cancer surgery without reconstruction when comparing prophylactic antibiotics to placebo. The relevant articles provide mixed results: some show a significant reduction in SSI with antibiotics (e.g., [7480237], [2403655]), while others show no significant difference (e.g., [23052809], [10873356], [2403697]). Notably, [7480237] specifically highlights a significant reduction in infections for breast surgery, while [2403655] also supports a reduction in infections for breast surgery. However, other studies, such as [23052809] and [10873356], found no significant difference. The variability in results suggests that the effect may depend on the type of breast surgery (e.g., mastectomy vs. axillary lymph node dissection) or other factors. 

- **Full Answer**: The incidence of SSI is **lower** with prophylactic antibiotics compared to placebo in some breast cancer surgeries, as shown by significant reductions in infections in studies involving plastic surgery ([7480237]) and certain breast procedures ([2403655]). However, other studies found no significant difference for modified radical mastectomy ([23052809]), clean elective breast surgery ([10873356]), or mastectomy ([2403697]). The overall evidence suggests a context-dependent effect, with some procedures benefiting from prophylaxis. 

- **Answer**: Lower
    \end{lstlisting}
    \noindent\rule{\linewidth}{0.4pt}
    \textbf{Remarks:} While DeepSeek V3 does get this question right, and shows flashes of attempting deeper reasoning (e.g. the remark that the effect "may depend on the type of breast surgery"), but this analysis is not directly supported by similar statements from the original SR authors. Additionally, there is still no formal aggregation or meta-analysis attempted, and the level of detail in the reasoning provided by DeepSeek V3 makes it hard to tell if the model's predicted answer was made via some form of intuition or if the model just got a bit lucky.
\end{QualExample}

\begin{QualExample}{teal}{Question ID 173 - 1/24 models correct}
    \textbf{Review PMID:} 31425604\\
    \textbf{Question:} Is the risk of necrotizing enterocolitis higher, lower, or the same when comparing monitoring of gastric residual volume and quality to only monitoring of gastric residual quality?\\
    \textbf{Correct Answer:} \texttt{uncertain effect}\\
    \textbf{PMIDs of Relevant Sources:} 29866595\\
    \noindent\rule{\linewidth}{0.4pt}\\
    \textbf{DeepSeek V3 Response:}
    \begin{lstlisting}
- **Rationale**: The study compared two groups: one where gastric residual volume (GRV) was measured only in the presence of specific clinical signs (study group) and another where GRV was routinely measured (control group). The primary outcome was time to reach full feeding volumes, but secondary outcomes included necrotizing enterocolitis (NEC). The study found no significant differences in NEC incidence between the two groups, with only 2 cases reported in the control group and none in the study group. However, the small sample size (87 infants) and low event rate limit the ability to draw definitive conclusions about NEC risk.

- **Full Answer**: The study found no significant difference in the incidence of necrotizing enterocolitis between the group where gastric residual volume was measured only in the presence of clinical signs (0 cases) and the group where it was routinely measured (2 cases) [29866595]. The small sample size and low event rate suggest uncertainty in the effect.

- **Answer**: Uncertain Effect
    \end{lstlisting}
    \noindent\rule{\linewidth}{0.4pt}
    \textbf{Remarks:} Even without explicit prompting, DeepSeek V3 recognizes the weakness of the limited sample size/total number of events—the fact that only DeepSeek V3 gets this question correct shows both the current limitations of models' ability to assess uncertainty, as well as the promise that they may be able to do so consistently in the future.
\end{QualExample}

\FloatBarrier

\section{Individual confusion matrices for all models}
\label{a:sec:allConfMats}
\begin{figure}
    \centering
    \includegraphics[width=0.6\linewidth]{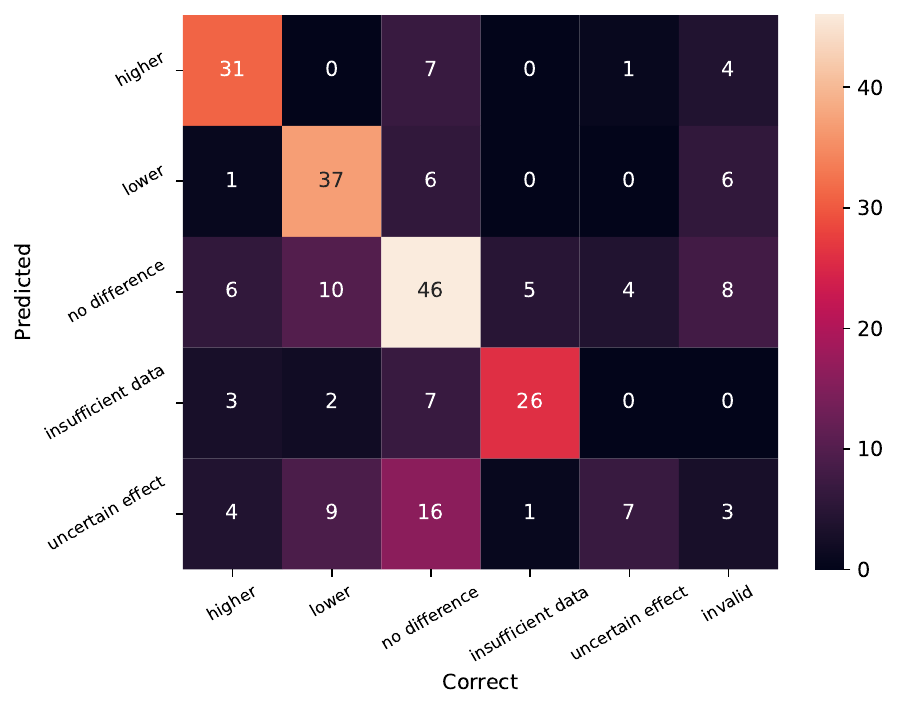}
    \caption{Confusion matrix for DeepSeek R1.}
\end{figure}

\begin{figure}
    \centering
    \includegraphics[width=0.6\linewidth]{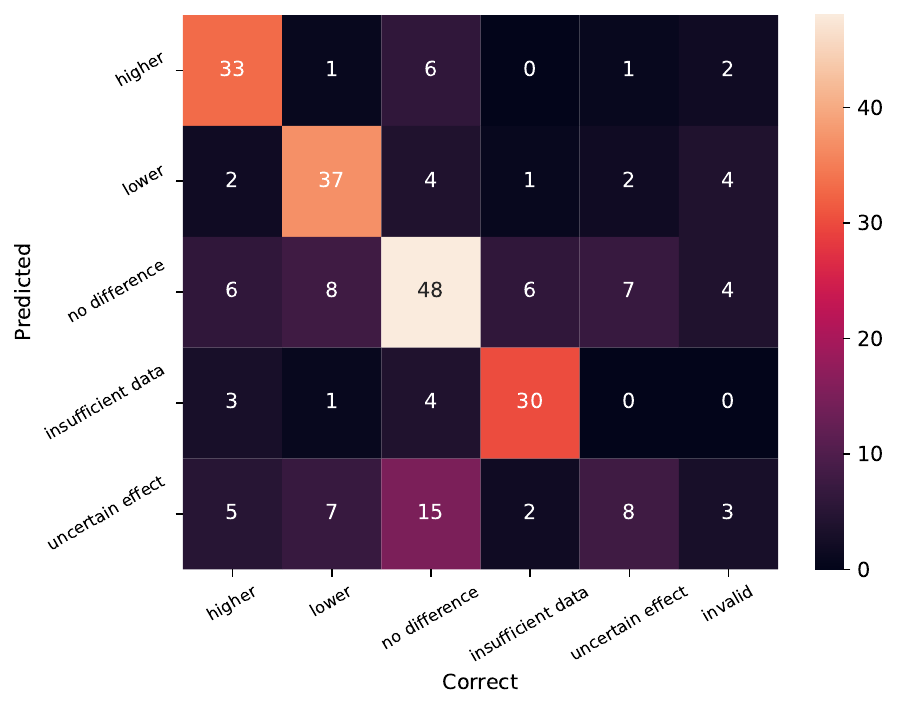}
    \caption{Confusion matrix for DeepSeek V3.}
\end{figure}

\begin{figure}
    \centering
    \includegraphics[width=0.6\linewidth]{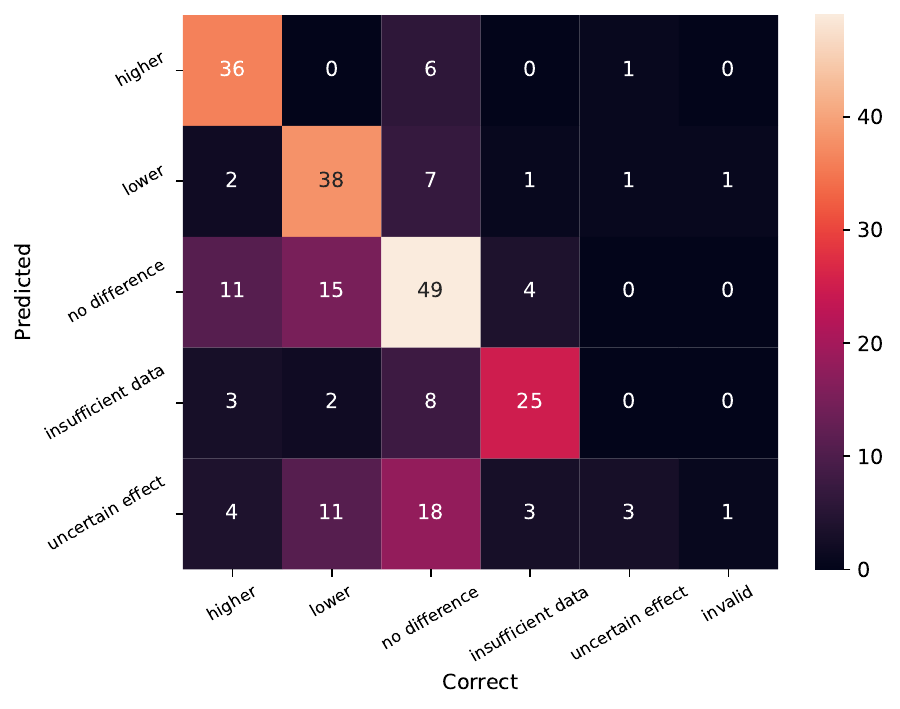}
    \caption{Confusion matrix for GPT-4.1.}
\end{figure}

\begin{figure}
    \centering
    \includegraphics[width=0.6\linewidth]{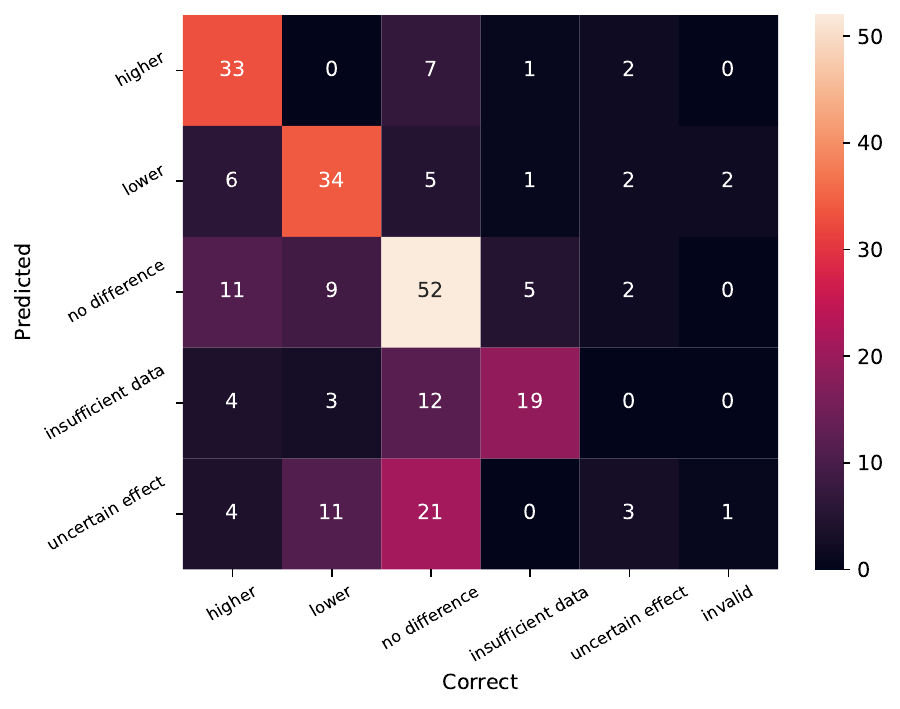}
    \caption{Confusion matrix for GPT-4.1 mini.}
\end{figure}

\begin{figure}
    \centering
    \includegraphics[width=0.6\linewidth]{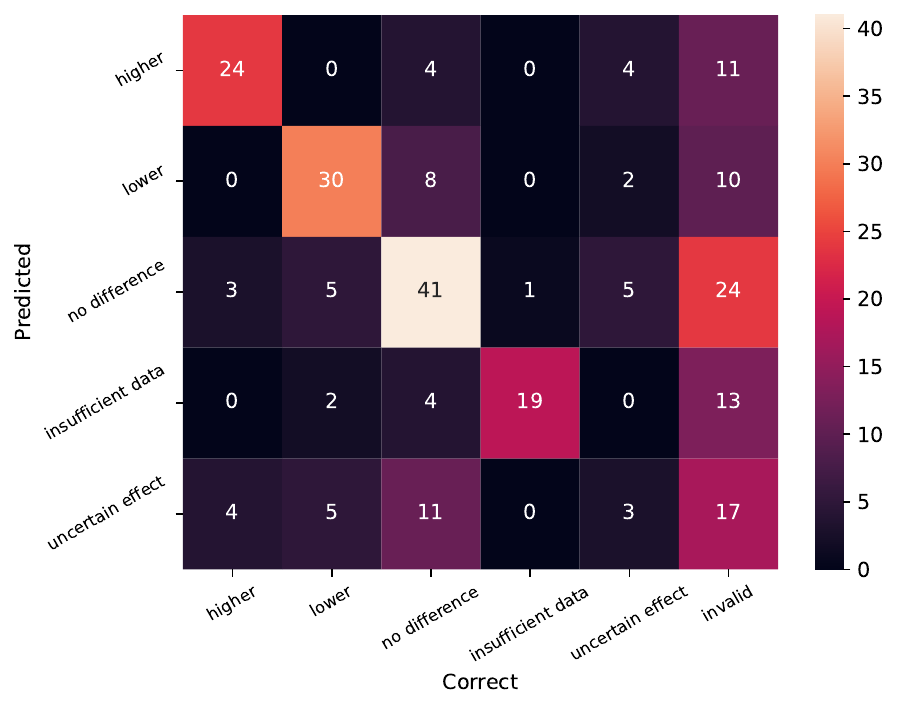}
    \caption{Confusion matrix for GPT-o1.}
\end{figure}

\begin{figure}
    \centering
    \includegraphics[width=0.6\linewidth]{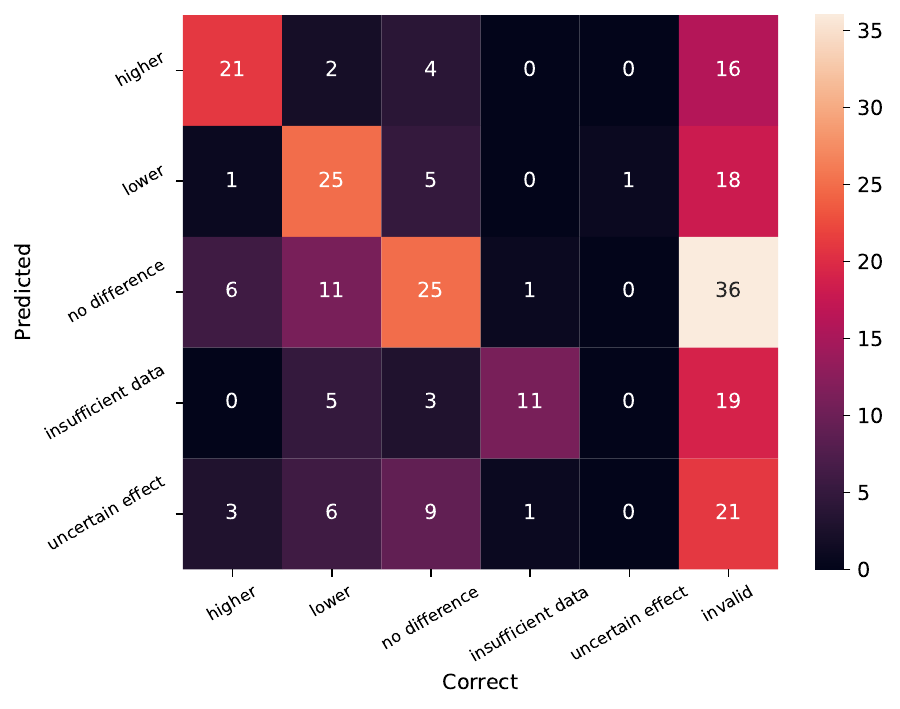}
    \caption{Confusion matrix for HuatuoGPT-o1-70B.}
\end{figure}

\begin{figure}
    \centering
    \includegraphics[width=0.6\linewidth]{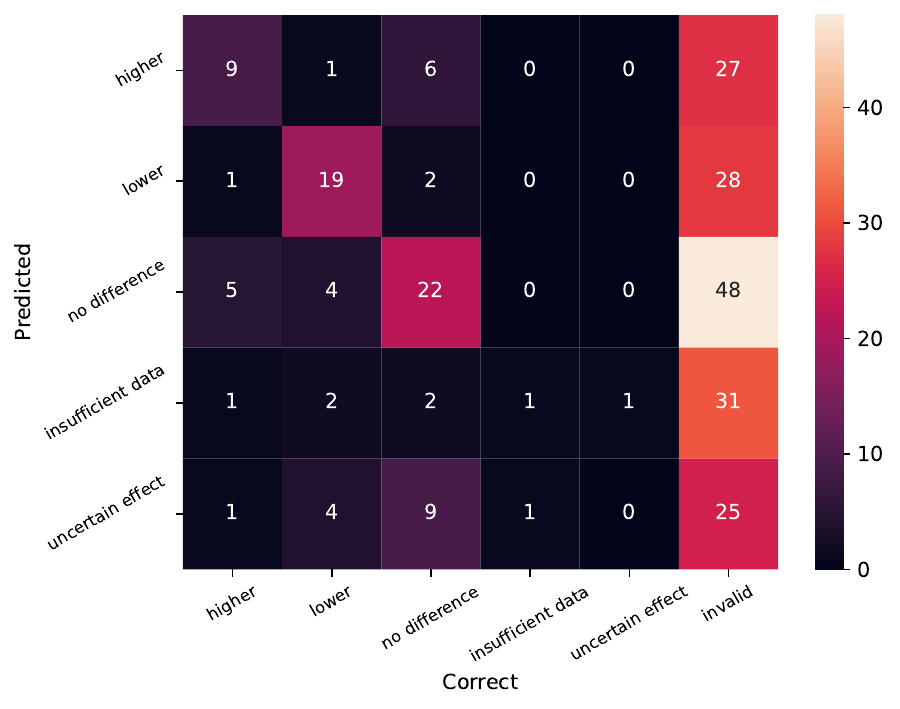}
    \caption{Confusion matrix for HuatuoGPT-o1-7B.}
\end{figure}

\begin{figure}
    \centering
    \includegraphics[width=0.6\linewidth]{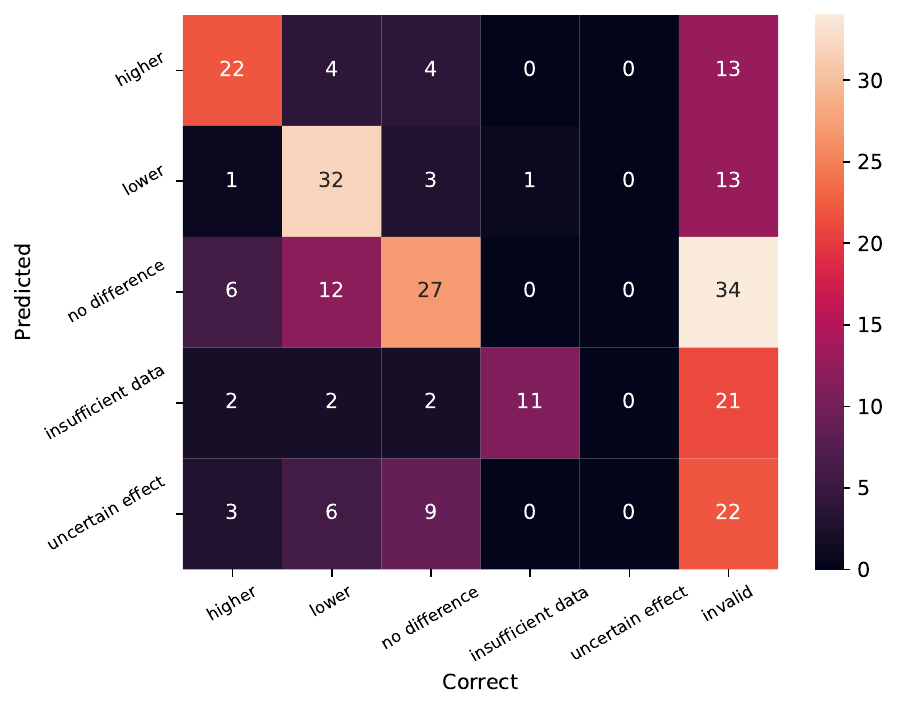}
    \caption{Confusion matrix for Llama 3.0 70B.}
\end{figure}

\begin{figure}
    \centering
    \includegraphics[width=0.6\linewidth]{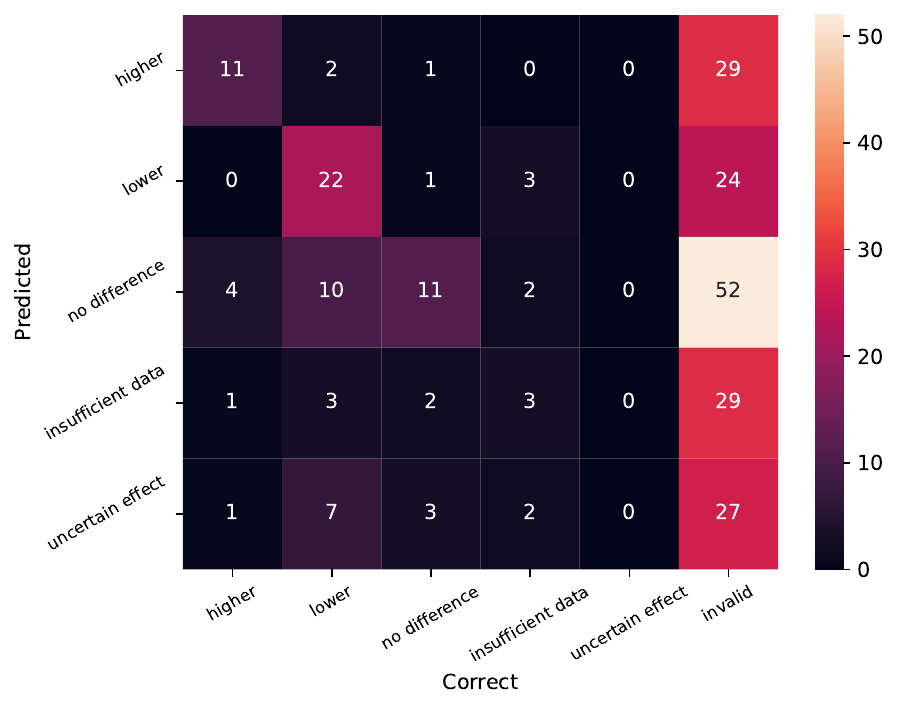}
    \caption{Confusion matrix for Llama 3.0 8B.}
\end{figure}

\begin{figure}
    \centering
    \includegraphics[width=0.6\linewidth]{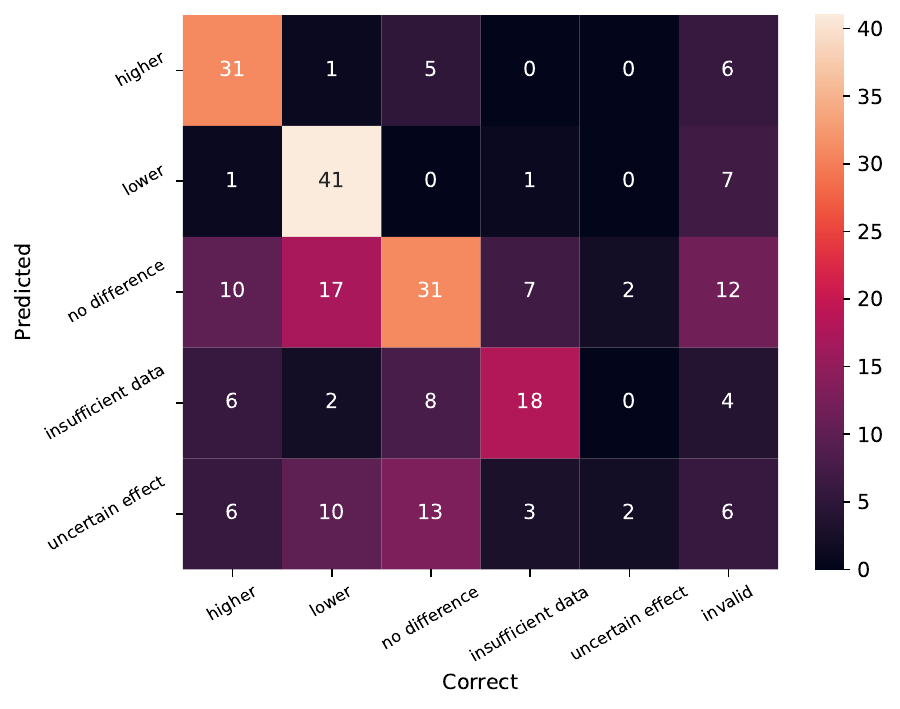}
    \caption{Confusion matrix for Llama 3.1 405B.}
\end{figure}

\begin{figure}
    \centering
    \includegraphics[width=0.6\linewidth]{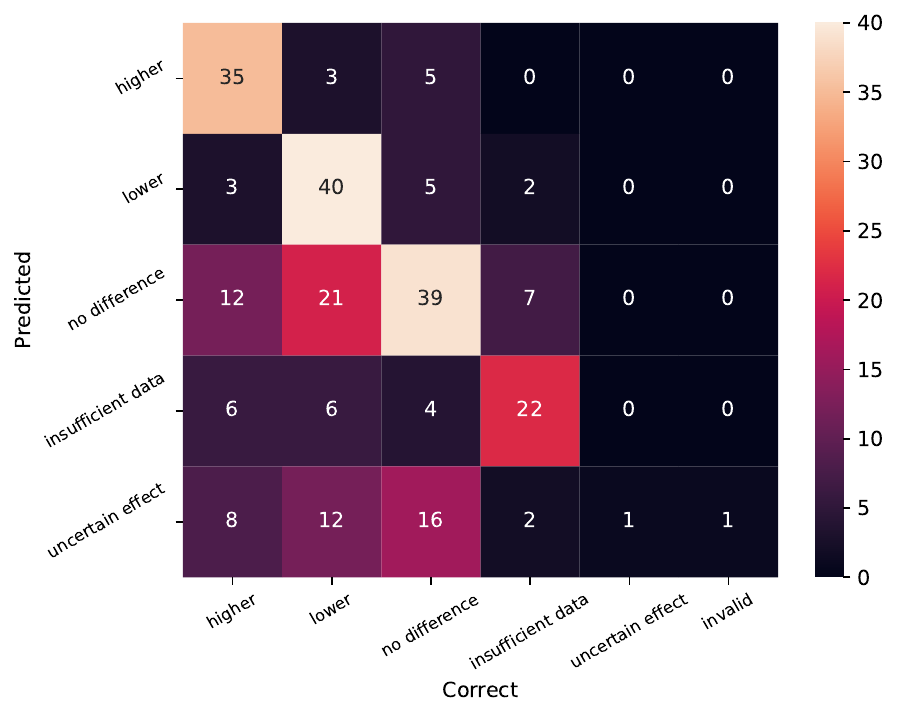}
    \caption{Confusion matrix for Llama 3.1 70B.}
\end{figure}

\begin{figure}
    \centering
    \includegraphics[width=0.6\linewidth]{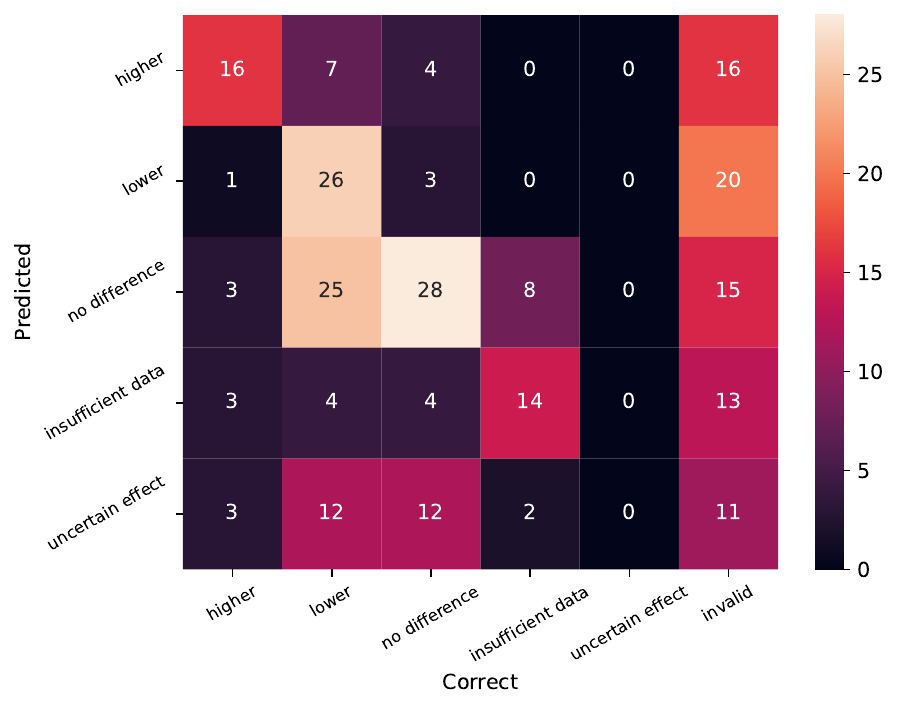}
    \caption{Confusion matrix for Llama 3.1 8B.}
\end{figure}

\begin{figure}
    \centering
    \includegraphics[width=0.6\linewidth]{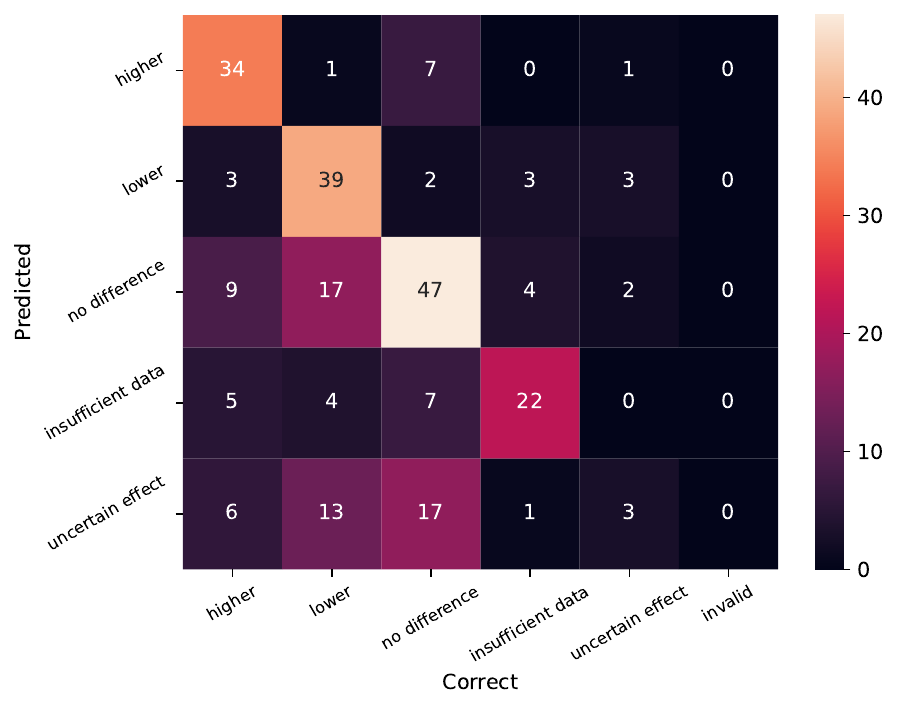}
    \caption{Confusion matrix for Llama 3.3 70B (R1-Distill).}
\end{figure}

\begin{figure}
    \centering
    \includegraphics[width=0.6\linewidth]{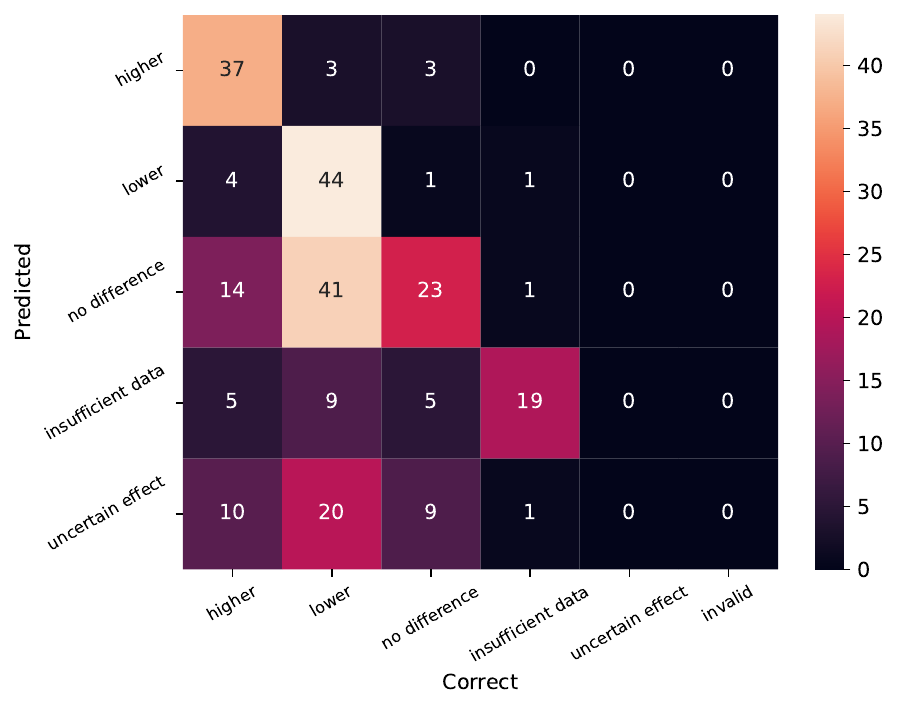}
    \caption{Confusion matrix for Llama 3.3 70B-Instruct.}
\end{figure}

\begin{figure}
    \centering
    \includegraphics[width=0.6\linewidth]{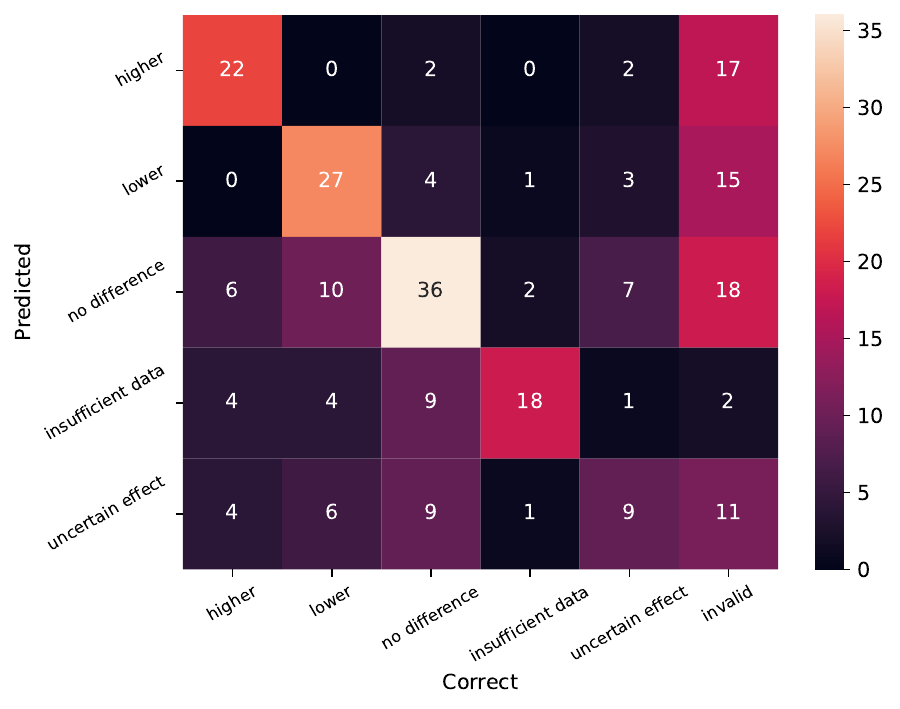}
    \caption{Confusion matrix for Llama 4 Maverick.}
\end{figure}

\begin{figure}
    \centering
    \includegraphics[width=0.6\linewidth]{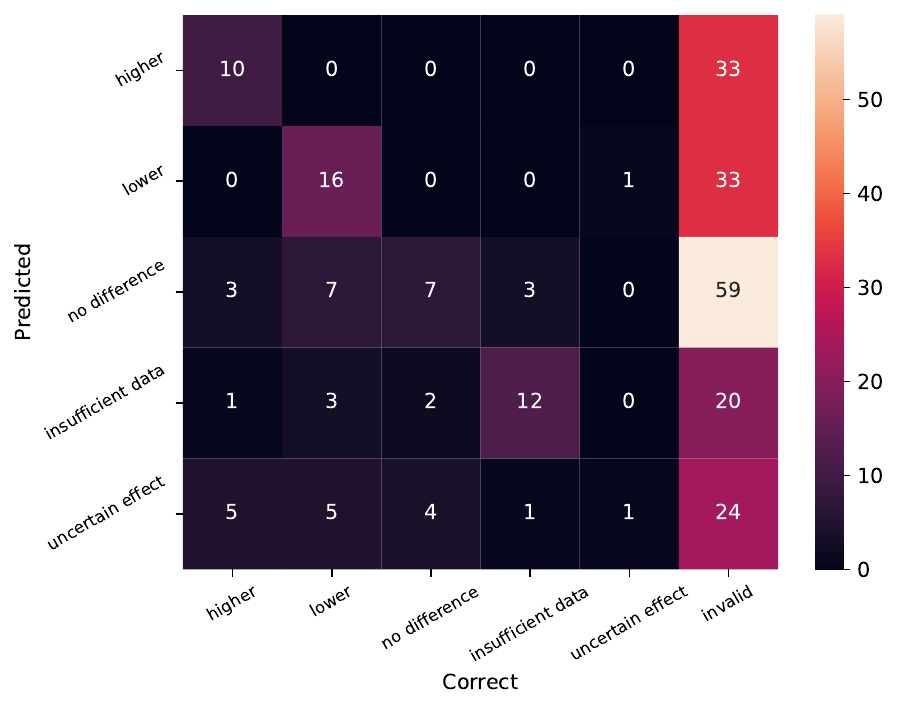}
    \caption{Confusion matrix for Llama 4 Scout.}
\end{figure}

\begin{figure}
    \centering
    \includegraphics[width=0.6\linewidth]{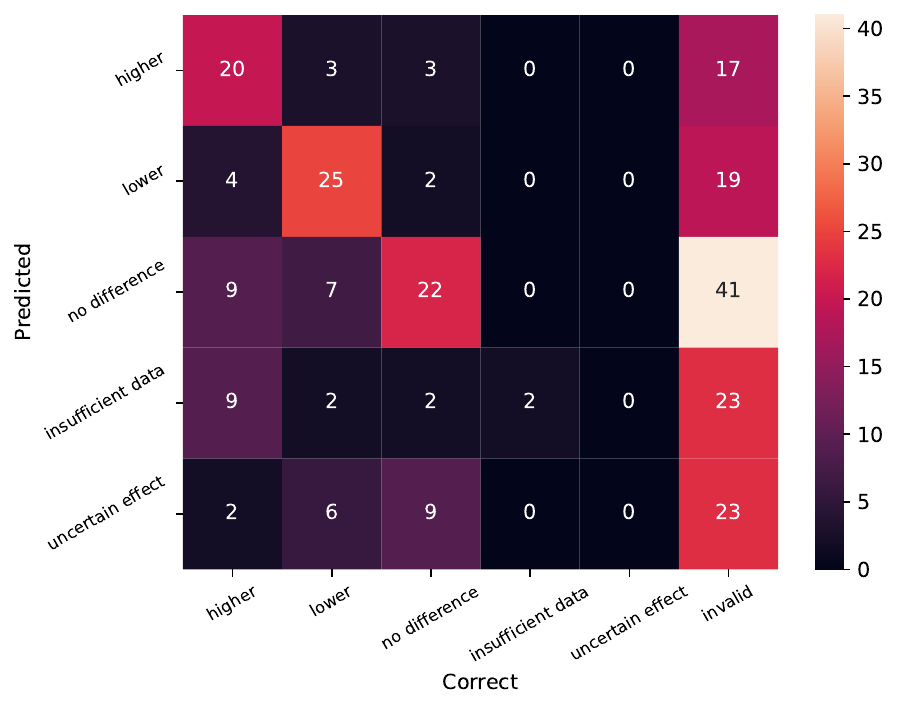}
    \caption{Confusion matrix for OpenBioLLM 70B.}
\end{figure}

\begin{figure}
    \centering
    \includegraphics[width=0.6\linewidth]{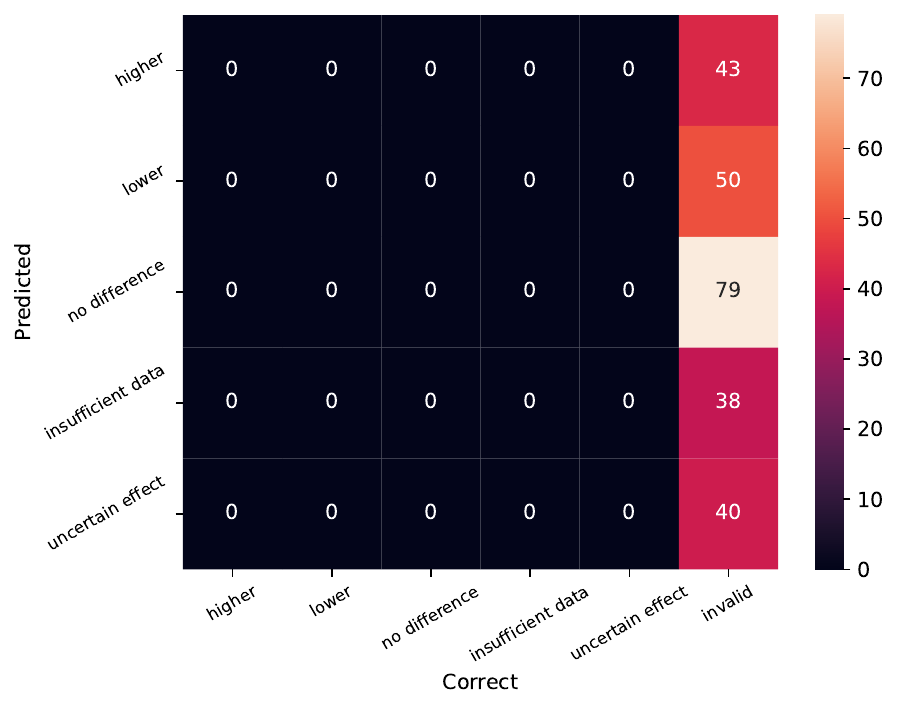}
    \caption{Confusion matrix for OpenBioLLM 8B.}
\end{figure}

\begin{figure}
    \centering
    \includegraphics[width=0.6\linewidth]{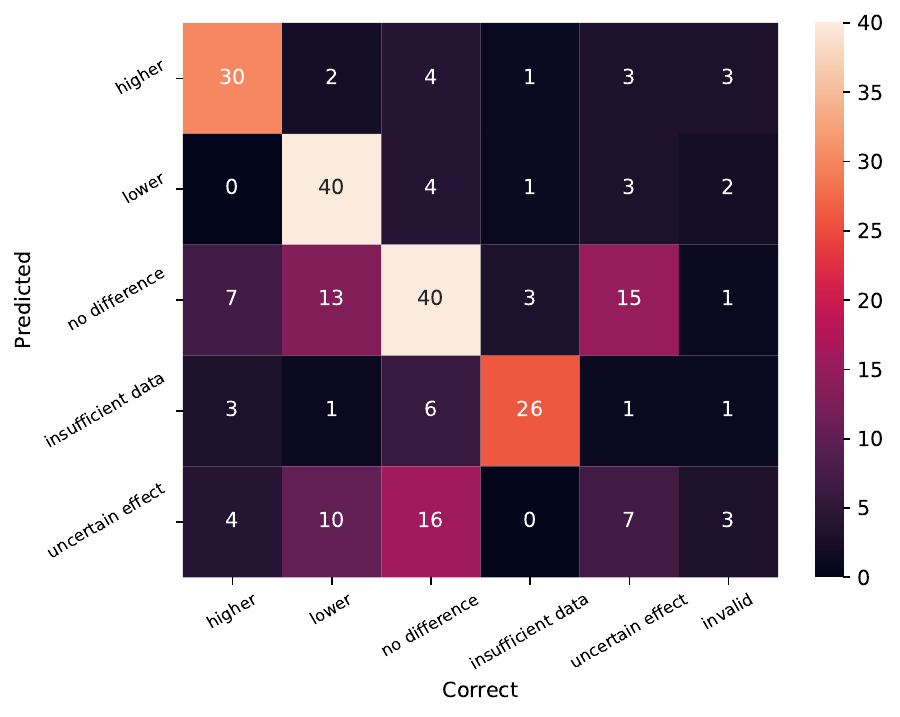}
    \caption{Confusion matrix for OpenThinker2-32B.}
\end{figure}

\begin{figure}
    \centering
    \includegraphics[width=0.6\linewidth]{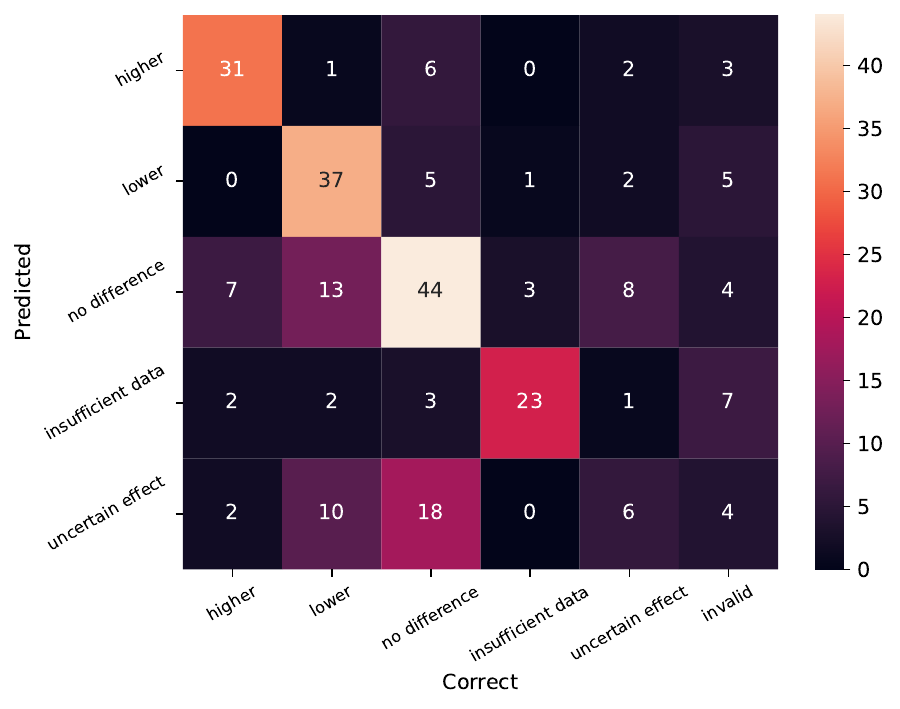}
    \caption{Confusion matrix for QwQ-32B.}
\end{figure}

\begin{figure}
    \centering
    \includegraphics[width=0.6\linewidth]{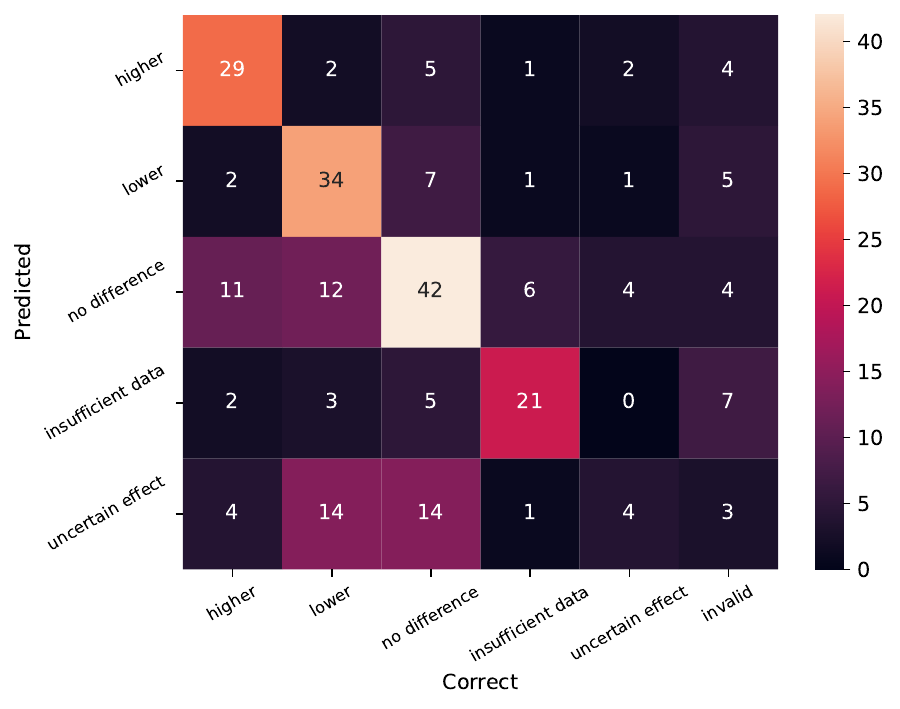}
    \caption{Confusion matrix for Qwen2.5-32B-Instruct.}
\end{figure}

\begin{figure}
    \centering
    \includegraphics[width=0.6\linewidth]{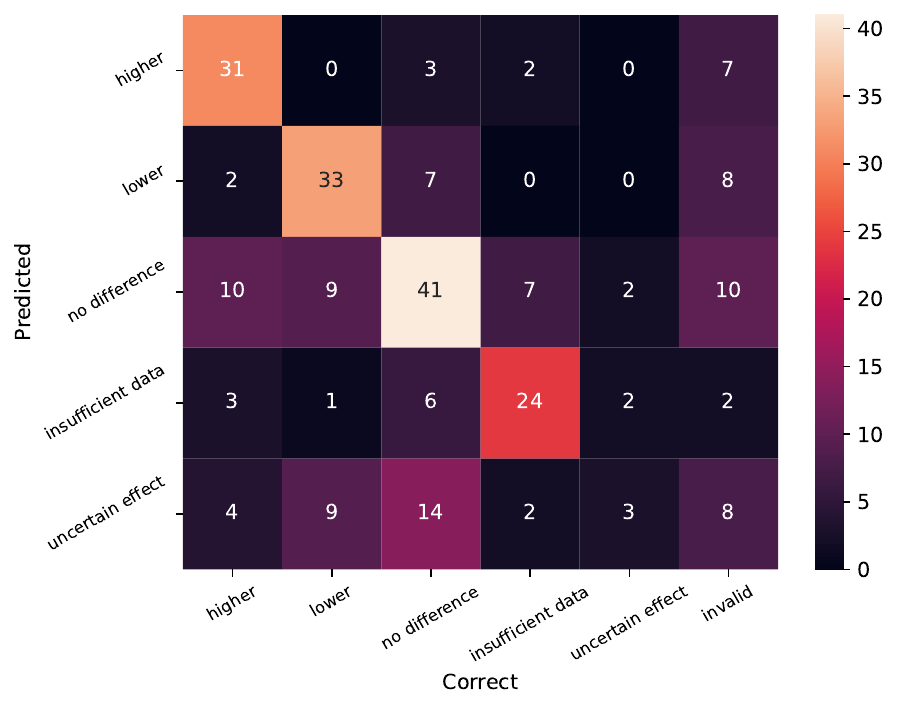}
    \caption{Confusion matrix for Qwen2.5-72B-Instruct.}
\end{figure}

\begin{figure}
    \centering
    \includegraphics[width=0.6\linewidth]{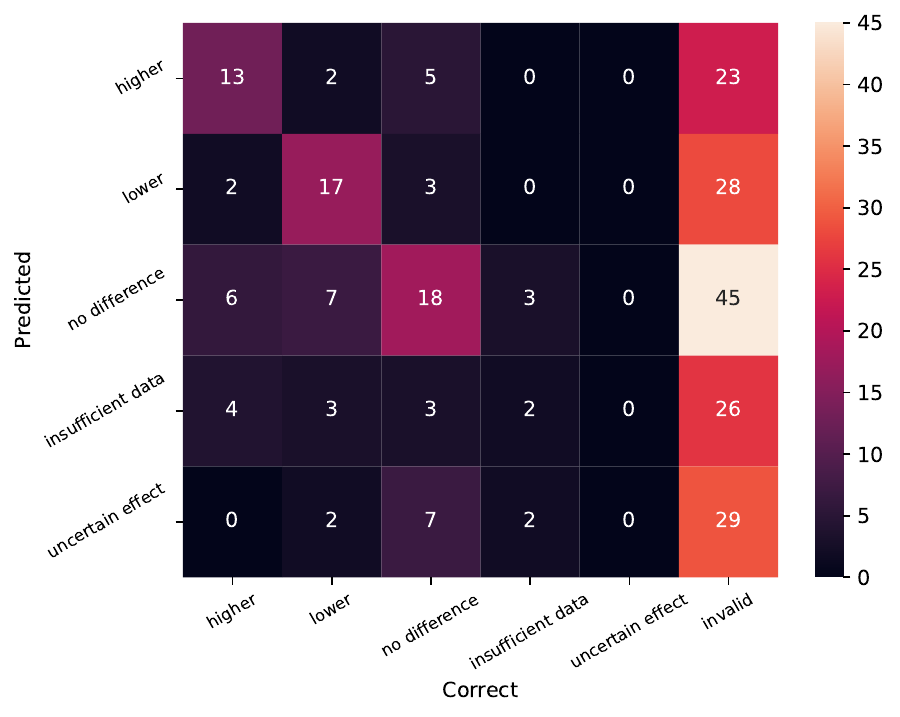}
    \caption{Confusion matrix for Qwen2.5-7B-Instruct.}
\end{figure}

\begin{figure}
    \centering
    \includegraphics[width=0.6\linewidth]{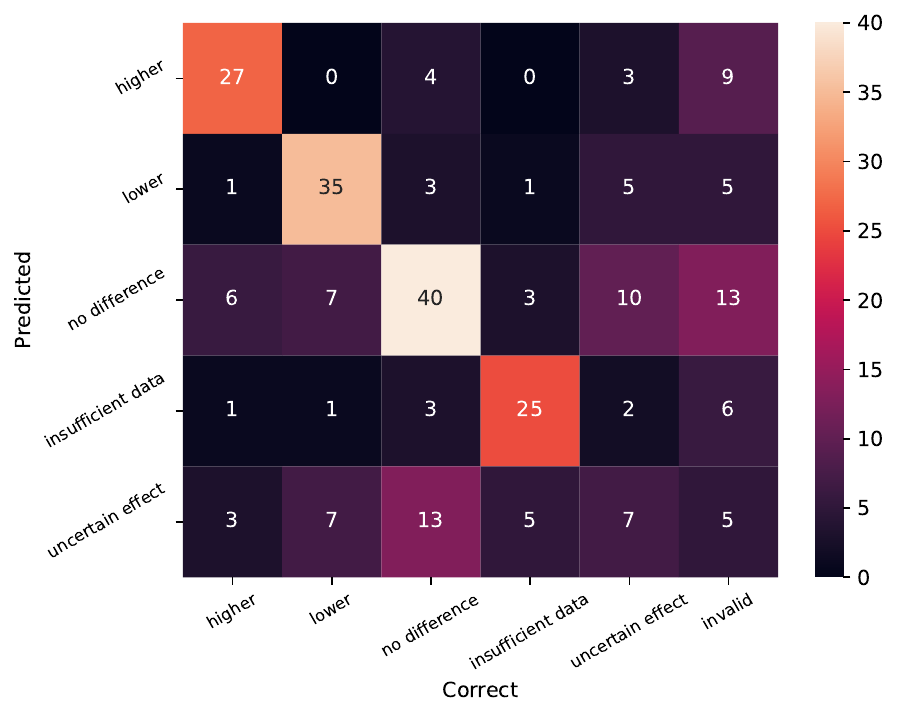}
    \caption{Confusion matrix for Qwen3-235B-A22B-FP8.}
\end{figure}

\end{document}